\PassOptionsToPackage{table}{xcolor}
\pdfoutput=1
\documentclass[11pt]{article}
\usepackage{float}
\usepackage{acl}
\usepackage[utf8]{inputenc}

\usepackage{pgfplots}
\usepackage{dsfont}

\DeclareUnicodeCharacter{2212}{−}
\usepgfplotslibrary{groupplots,dateplot}
\usetikzlibrary{patterns,shapes.arrows}
\pgfplotsset{compat=newest}
\pgfplotsset{width=10cm,compat=1.9}

\usepackage{tikzscale} 
\usepackage[utf8]{inputenc}
\usepackage{bbm}
\newcommand{\probP}{\text{I\kern-0.15em P}}

\usepackage{colortbl} 
\usepackage{makecell}

\usepackage{graphicx}

\usepackage{multirow}
\usepackage{array}
\usepackage{amssymb}
\usepackage{pifont}
\usepackage{CJKutf8}
\usepackage{tabularx}
\usepackage{placeins}
\usepackage[normalem]{ulem}
\usepackage{booktabs}

\usepackage[]{algpseudocode}
\usepackage[]{algorithm}
\algtext*{EndFor}%
\algtext*{EndProcedure}%
\usepackage[normalem]{ulem}
\usepackage{times}
\usepackage{latexsym}
\usepackage{adjustbox}

\usepackage[T1]{fontenc}

\usepackage{amsmath}
\usepackage{scalerel,xparse}

\usepackage{cleveref}
\usepackage{microtype}

\usepackage[normalem]{ulem}
\usepackage{fontawesome5}
\usepackage{color}
\usepackage{xspace}
\usepackage{todonotes}
\usepackage{enumitem}
\usepackage[most]{tcolorbox}
\usepackage{subfigure}

\definecolor{bggray}{rgb}{0.95, 0.95, 0.95}
\usepackage[%
    framemethod=tikz,
    skipbelow=\topskip,
    skipabove=\topskip
]{mdframed}
\mdfsetup{%
    leftmargin=0pt,
    rightmargin=0pt,
    backgroundcolor=bggray,
    middlelinecolor=black,
    roundcorner=3
}
\newtcolorbox[list inside=prompt,auto counter,number within=section]{prompt}[1][]{
    colbacktitle=black!60,
    fonttitle=\small,
    coltitle=white,
    fontupper=\footnotesize,
    boxsep=4pt,
    left=0pt,
    top=0pt,
    bottom=0pt,
    boxrule=1pt,
    #1,
}

\usepackage[T1]{fontenc}
\usepackage[russian,english]{babel}
\usepackage{tikz}

\usepackage{soul}
\sethlcolor{orange!50}

\definecolor{zoey green}{rgb}{0.684,0.836,0.227}

\newcommand{\ignore}[1]{}

\title{Should I Share this Translation? \\
Evaluating Quality Feedback for User Reliance on Machine Translation}

\author{\textbf{Dayeon Ki}\textsuperscript{\ding{68}} \hspace{0.5cm} 
        \textbf{Kevin Duh}\textsuperscript{\ding{83}} \hspace{0.5cm} 
        \textbf{Marine Carpuat}\textsuperscript{\ding{68}} \vspace{0.1cm}  \\
  \textsuperscript{\ding{68}}University of Maryland \hspace{0.5cm}
  \textsuperscript{\ding{83}}Johns Hopkins University \hspace{0.5cm} \\
  \texttt{\{dayeonki,marine\}@umd.edu}
  \hspace{0.5cm}
  \texttt{kevinduh@cs.jhu.edu}
}

\begin{document}
\maketitle

\begin{abstract}



As people increasingly use AI systems in work and daily life, mechanisms that help them use AI responsibly are urgently needed, especially when they are not equipped to verify AI predictions themselves. We study a realistic Machine Translation (MT) scenario where monolingual users decide whether to share an MT output, first without and then with quality feedback. We compare four types of quality feedback: explicit feedback that directly give users an assessment of translation quality using (1) error highlights and (2) LLM explanations, and implicit feedback that helps users compare MT inputs and outputs through (3) backtranslation and (4) question–answer (QA) tables. We find that all feedback types, except error highlights, significantly improve both decision accuracy and appropriate reliance. Notably, implicit feedback, especially QA tables, yields significantly greater gains than explicit feedback in terms of decision accuracy, appropriate reliance, and user perceptions \---\ receiving the highest ratings for helpfulness and trust, and the lowest for mental burden.\footnote{\url{https://github.com/dayeonki/mt_quality_feedback}}

\end{abstract}

\section{Introduction}

\definecolor{lightpink}{RGB}{255, 237, 250}
\definecolor{midpink}{RGB}{225, 149, 171}
\definecolor{darkpink}{RGB}{222, 49, 99}
\definecolor{lime}{RGB}{204, 223, 146}
\definecolor{skyblue}{RGB}{179, 218, 253}

\begin{figure*}
    \centering
    \includegraphics[width=\linewidth]{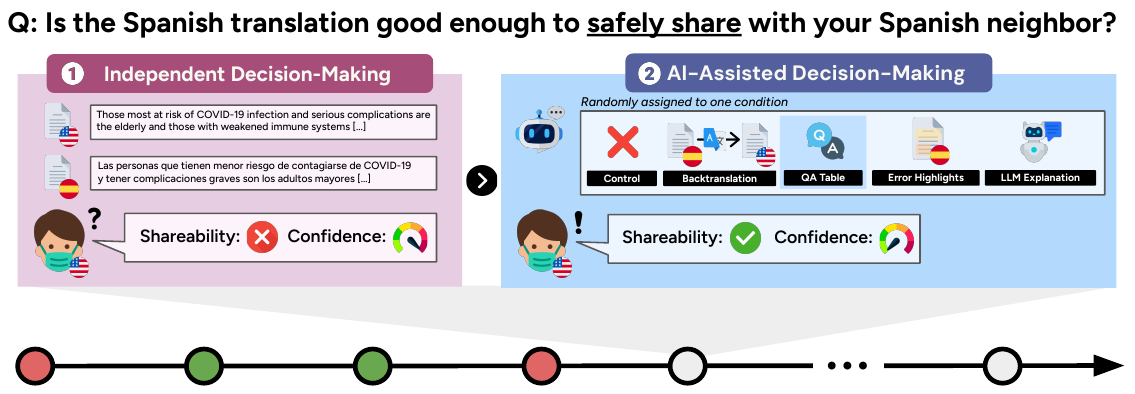}
    \caption{\textbf{Overview of our study setup.} In our human study, each English-speaking monolingual participant reviews a sequence of 20 decision-making examples. Each example is shown in a two-step process: \textbf{\ding{202} Independent decision-making:} Participants first make judgments based solely on the English source and its Spanish MT output \colorbox{midpink}{  } and \textbf{\ding{203} AI-Assisted decision-making:} They then reassess the same example with one of five randomly assigned conditions (one control and four treatments) \colorbox{skyblue}{  }. For each step, they respond to two questions: \textbf{(1) Shareability:} To the best of your knowledge, is the Spanish translation good enough to safely share with your Spanish-speaking neighbor? and \textbf{(2) Confidence:} How confident are you in your assessment?}
    \label{fig:main_figure}
\end{figure*}


Artificial Intelligence (AI) systems are increasingly deployed to support human decision-making across a wide range of domains \citep{to_trust_or_to_think, dastin2022amazon, ma2024recommenderexploratorystudyeffects}. As these systems are adopted by the general public, there is a growing need for feedback mechanisms that help users construct their own functional explanations \---\ reasoning grounded in the goals and consequences of an AI output to determine how and when to rely on it for safe and effective use \citep{mechanistic_functional, Schoeffer_2024}. Prior work has evaluated various forms of feedback through human studies, such as error highlights \citep{khashabi2018looking, carton2020evaluating} or free-text explanations \citep{bussone2015role, bansal2021does, to_trust_or_to_think}. 
However, many human-centered studies evaluating the impact of feedback in real-world application settings are still needed. Although designing such studies is challenging, as it requires accounting for the knowledge and assumptions people bring to decision-making tasks \citep{lage2019evaluationhumaninterpretabilityexplanation, wiegreffe2021teach}, it remains essential to adopt a human-centered evaluation since these feedback mechanisms are ultimately intended to support human users \citep{boyd-graber-etal-2022-human, zhu-etal-2024-explanation, carpuat2025interdisciplinaryapproachhumancenteredmachine}.


To address this, and in line with prior work that empirically investigates real-world use cases in other domains \citep{sungsoo_2020_humanfactors, umang_2020_explainable, vera_liao_2020}, we evaluate AI feedback for Machine Translation (MT), where many users critically need support because they lack the language proficiency needed to evaluate MT outputs.
Imagine a situation during the COVID-19 pandemic: You regularly read official guidelines in English, but your Spanish-speaking neighbor cannot access this information. You turn to MT to share this information \---\ but as an English monolingual, how can you determine whether the Spanish MT output is accurate enough to safely share, or if it contains critical errors risking misinformation? 
This is a practical yet challenging scenario for monolingual users, who lack both source language proficiency \citep{bowker2019translation, liebling_2020} and domain expertise \citep{nourani2020role, lee2023understanding} to reliably evaluate MT quality, and who often lack effective strategies for deciding when to trust imperfect MT \citep{beyondbenchmarks_2025}.


Prior studies have proposed quality feedback mechanisms to support MT decision-making, such as paraphrases, Quality Estimation (QE) scores, and backtranslation, but findings on their impact on user confidence and decision accuracy remain mixed \citep{zouhar2021backtranslation, mehandru-etal-2023-physician}. Building on this, we provide a more comprehensive assessment of which feedback types best support users in forming functional explanations to make reliable MT decisions. We evaluate four types of quality feedback, grouped by their mode of explicitness:
\textbf{(1) Explicit} quality assessments of MT output (error highlights and LLM explanation); and
\textbf{(2) Implicit} assessments that support input/output comparison (backtranslation and QA table).

We conduct a between-subjects human study with 91 English-speaking monolingual participants, where they are asked to decide whether Spanish MT outputs are safe to share with a hypothetical Spanish-speaking neighbor. For each of 20 examples, participants first make a decision (Independent step), then reassess the same example with a randomly assigned condition (AI-Assisted step), as illustrated in Figure~\ref{fig:main_figure}.

Our findings reveal that all quality feedback interventions except error highlights significantly improve both decision accuracy and appropriate reliance (\S\ref{sec:results_rq1}). Implicit feedback generally outperforms explicit types: backtranslation yields significantly higher appropriate reliance than error highlights, and QA table leads to significantly higher gains in both metrics than both explicit feedback types (\S\ref{sec:results_rq2}). While explicit feedback prompts more decision changes, it also results in higher over-reliance, which are cases where participants change from correct to incorrect decisions after viewing feedback (\S\ref{sec:results_rq3}). We also find that participants are better at recognizing good translations than identifying problematic ones (\S\ref{sec:shareability}). In terms of self-reported perceptions on mental burden, helpfulness, and trust, QA table feedback consistently receives the best ratings (\S\ref{sec:perception}). Finally, we present participants' responses to identify which aspects of each quality feedback they found to be helpful (\S\ref{sec:free_comments}).

Together, these results highlight the value of quality feedback that supports users’ implicit interpretation of MT outputs rather than explicitly telling them \textit{what} to do, supporting a more \textit{user-driven} process for making reliable decisions.

\section{Background \& Research Questions}

\subsection{Reliance on AI Systems}
\label{sec:2.1}


A growing body of work has examined the nature of human reliance on AI systems \citep{lai2023towards}, particularly in scenarios involving risk and uncertainty \citep{jacovi2021formalizing}. While the overarching goal is to design trustworthy AI, studying trust is complex and multifaceted. To operationalize this, prior works have developed methods to study human behavior when using AI systems with a focus on reliance \citep{de2021defining}, defined as the ``decision to follow someone's recommendation'' \citep{vereschak_2021}. Various metrics have been proposed to assess the degree of user reliance, including agreement percentage (how often a user agrees with the AI prediction) \citep{gaole_2023}, confidence-weighted accuracy (decision accuracy weighted by user confidence) \citep{mehandru-etal-2023-physician}, and switch percentage (how often a user changes their decision after seeing AI feedback) \citep{schmitt2021trust}.


One of the core challenges in human-AI collaboration is achieving appropriate reliance, which is accepting correct AI advice while rejecting incorrect advice \citep{eckhardt2024surveyaireliance}. In contrast, under-reliance (rejecting correct advice) and over-reliance (accepting incorrect advice) are both undesirable. We build our work on this line of measuring user reliance in AI systems, but more specifically in the context of Machine Translation (MT).

\subsection{Impact of Feedback on Reliance}
\label{sec:2.2}

In AI-assisted decision-making, AI systems typically play a supportive role, offering explanations in various formats, such as recommendations, confidence scores \citep{yin2019understanding}, (un)certainty estimates, output rationales \citep{bussone2015role, bansal2021does, to_trust_or_to_think}, or a combination thereof \citep{zhang2020effect}. These AI feedback are intended to help human users decide whether and when to rely on AI outputs \citep{lai2021towards}. However, empirical studies have shown that despite the intended benefits of AI explanations in fostering human–AI collaboration, they often lead to increased user confidence without corresponding improvements in decision accuracy, resulting in over-reliance on the AI system \citep{bansal2021does, poursabzi2021manipulating, sunnie_2025}.

\subsection{Impact of Quality Feedback on Reliance on MT}
\label{sec:2.3}

In the context of MT, the role of the AI system (i.e., MT system) takes on a different character since monolingual users often lack the mechanisms to reliably assess MT quality, and the AI prediction (i.e., MT output) is not a direct prediction for the user’s decision-making task. This unique property allows MT systems to be used for many \textit{implicit} decision-making scenarios (\textit{e.g.,} Is this translation good enough to share with a friend? To translate an official document?). This contrasts to traditional AI-assisted decision-making tasks, which typically focus on classification settings (\textit{e.g.,} recidivism prediction \citep{wang_2021}), where the AI prediction directly maps onto a single decision. 
Given this difference, our work focuses on quality feedback, which are more generic assessments of MT quality rather than direct recommendations, and ask whether users can rely on such feedback to make more informed decisions.

Various forms of quality feedback have been proposed, including quality estimation (QE) score, backtranslation \citep{agrawal-etal-2022-quality}, error highlights that flag problematic spans in the MT output \citep{eksi-etal-2021-explaining, rubino-etal-2021-error, briakou-etal-2023-explaining}, textual explanations of metric outputs \citep{fomicheva-etal-2022-translation, xu-etal-2023-instructscore, jiang2024tigerscore, lu-etal-2024-error}, and question–answer (QA) pairs designed to indicate potential errors in the translation \citep{sugiyama-etal-2015-investigation, krubinski-etal-2021-just, han-etal-2022-simqa, ki2025askqequestionansweringautomatic, fernandes2025llmsunderstandtranslationsevaluating}.

Only a few human studies have evaluated quality feedback on user decision-making and reliance on MT. For example, \citet{zouhar2021backtranslation} show that backtranslation significantly increased user confidence in translations, even when it did not improve decision accuracy. \citet{mehandru-etal-2023-physician} demonstrate that backtranslation can help users detect critical errors more effectively than QE scores in clinical settings. However, results across these studies remain mixed, and comparisons with more recent feedback mechanisms are still lacking. Our work aims to address this gap.

\paragraph{Research Questions.} Given this context, we address the following RQs:
\begin{itemize}[leftmargin=*, itemsep=2pt, parsep=-1pt]
    \item[] \textbf{RQ1.} How accurately and appropriately do monolingual users decide whether to share translations when provided with quality feedback?
    \item[] \textbf{RQ2.} How does their decision-making performance vary across different quality feedback and the two modes of explicitness?
    \item[] \textbf{RQ3.} How do users change their decisions in response to each type of quality feedback?
\end{itemize}

\section{Methods}

In this section, we describe the experimental study conducted to address our RQs. We outline the overall study design (\S\ref{sec:study_design}), four types of quality feedback used as treatment conditions (\S\ref{sec:feedback_type}), stimuli collection process (\S\ref{sec:stimuli_collection}), participant details (\S\ref{sec:demographics}), and our dependent variables (\S\ref{sec:evaluation}).

\definecolor{qablue}{RGB}{57, 133, 219}

\begin{figure*}
    \centering
    \includegraphics[width=\linewidth]{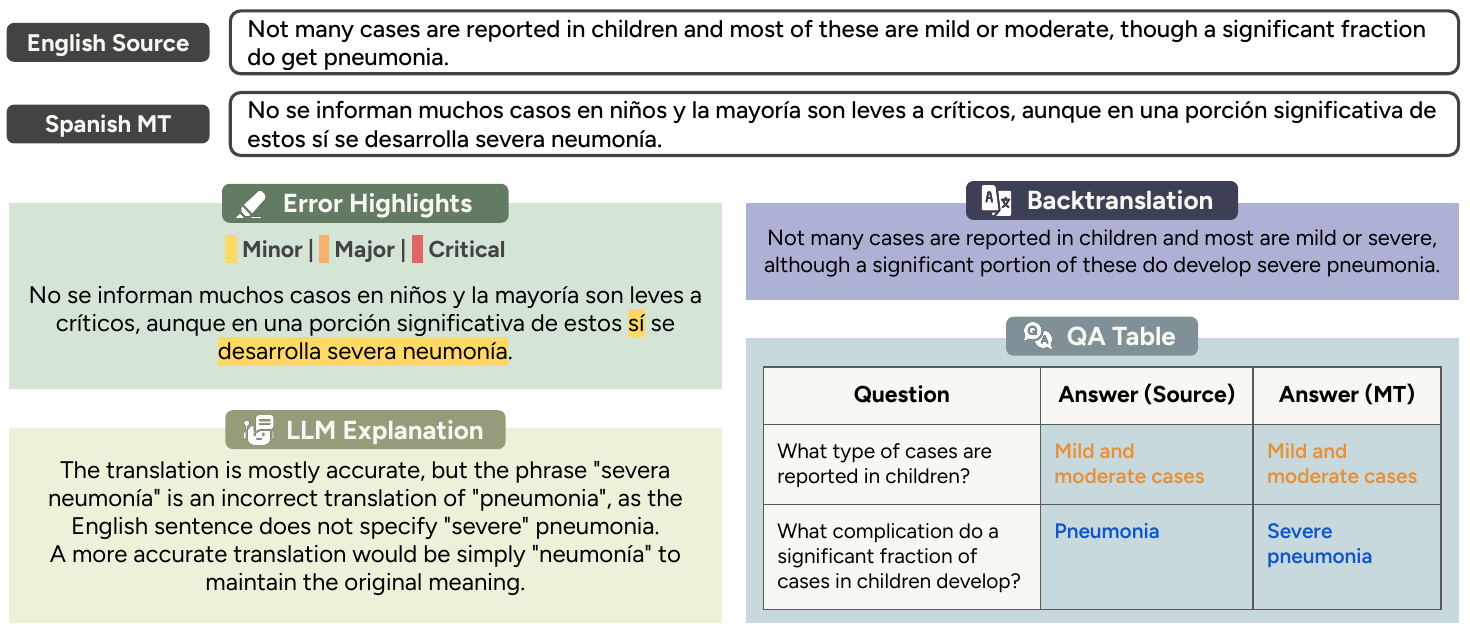}
    \caption{During the AI-assisted decision-making step, each \textit{treatment} group participant is presented with an English source, Spanish translation, and one of four randomly assigned quality feedback types. For error highlights, we also show a color-coded legend (\colorbox{yellow}{  } Minor | \colorbox{orange!60}{  } Major | \colorbox{red!50}{  } Critical) and for QA table, answer texts are displayed in \textcolor{orange}{orange} when they are identical or highly similar, else, \textcolor{qablue}{blue}.}
    \label{fig:feedback_types}
\end{figure*}

\subsection{Study Design}
\label{sec:study_design}
We study how different types of quality feedback impact users’ decision-making regarding MT shareability through a sequence of 20 examples in a between-subjects design, as illustrated in Figure~\ref{fig:main_figure}. We use the notion of shareability to capture not only perceived MT quality but also the potential \textit{risk} of miscommunication, highlighting the potential consequences in high-stakes contexts. This framing aligns with how people make such decisions in practice: while the choice to share or not is often made implicitly in real-world use cases, our study makes this decision more explicit, yet still allows participants to make their own judgment as they naturally would.

Specifically, we situate participants in a scenario where an English monolingual speaker reads official COVID-19 guidelines in English and decide whether the Spanish MT output is of sufficient quality to safely share with a Spanish-speaking neighbor. Each example is presented in two steps: \ding{202} \textbf{Independent} step, where participants first make judgments based solely on the English source and its Spanish translation, and a subsequent \ding{203} \textbf{AI-Assisted} step, where the same example is either shown again (\textit{control} condition) or paired with a specific type of quality feedback to support decision-making (\textit{treatment} condition). Examples are presented in randomized order, and two attention checks were included.


\paragraph{\ding{202} Independent Decision-Making.}
For each example, participants are first asked to make judgments based solely on an English sentence and its corresponding Spanish translation. They are asked to answer two questions: \textbf{(1) Shareability:} To the best of your knowledge, is the Spanish translation good enough to safely share with your Spanish-speaking neighbor? (with binary options: \raisebox{-0.2em}{\includegraphics[height=1.1em]{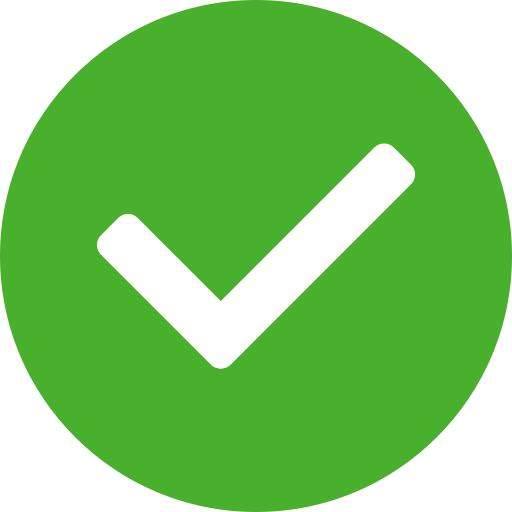}} Safe to share as-is, and \raisebox{-0.2em}{\includegraphics[height=1.1em]{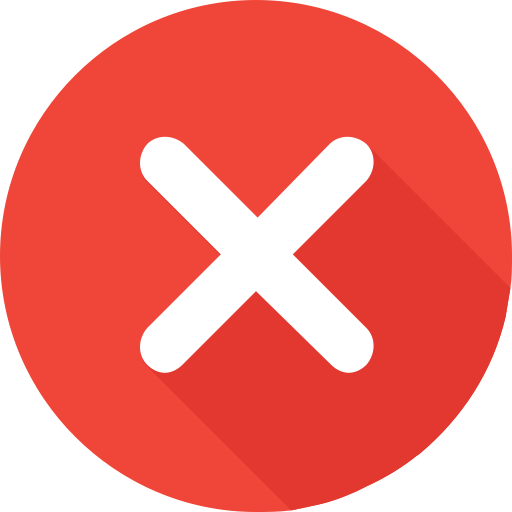}} Needs bilingual review before sharing); and \textbf{(2) Confidence:} How confident are you in your assessment? (on a five-point Likert scale from 1:Very Unconfident to 5:Very Confident). Since the recruited participants are monolingual English speakers, they are instructed to do their best in assessing the shareability of the MT outputs, despite not being fluent in Spanish.

\paragraph{\ding{203} AI-Assisted Decision-Making.}
Subsequently, participants are randomly assigned to one of five conditions: a control condition or one of four treatment conditions, each involving a different type of quality feedback (\S \ref{sec:feedback_type}). Those in the control condition view the same 20 examples twice in succession without receiving any quality feedback. In all conditions, participants answer the same two questions: shareability and confidence.



\paragraph{Pre-/Post-Task Survey.}
Before starting the main study, each participant is asked to answer four pre-task questions regarding their first language, proficiency in English and Spanish, and frequency of using AI translation tools in daily life or work. After completing the main study, each participant answers three post-task questions about their experience with the randomly assigned condition in terms of perceived mental burden, helpfulness, and trust for future use \citep{hoffman2019metricsexplainableaichallenges}. Detailed descriptions are provided in Appendix \ref{appendix:annotation_interface}.

\subsection{Types of Quality Feedback Intervention}
\label{sec:feedback_type}
Figure~\ref{fig:feedback_types} illustrates an example of each of four quality feedback interventions. Error highlights and LLM explanation provide \textbf{explicit} quality assessments of MT output, whereas backtranslation and QA table offer \textbf{implicit} assessments that guide participants to compare MT input and output. Details on how each feedback is shown to participants are in Appendix~\ref{appendix:annotation_interface}. To control for feedback quality, we balance the error rates of feedback predictions across all types (Appendix~\ref{appendix:error_rate}).\footnote{We compare the computational and time efficiency for generating each feedback in Appendix~\ref{appendix:efficiency}.}

\paragraph{\raisebox{-0.2em}{\includegraphics[height=1.1em]{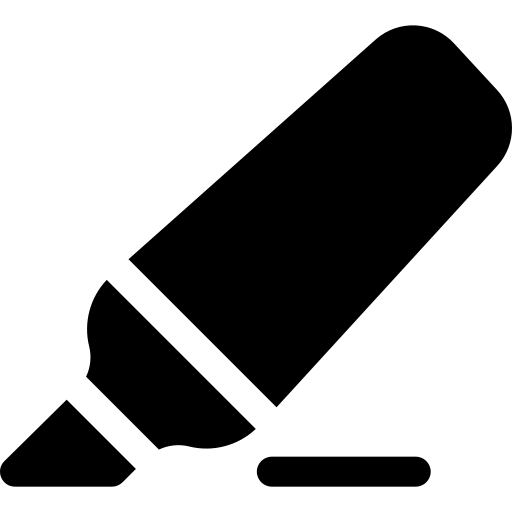}} Error Highlights.}
We adopt an off-the-shelf QE system, \textsc{xCOMET-XXL}\footnote{\url{https://huggingface.co/Unbabel/XCOMET-XXL}} \citep{guerreiro-etal-2024-xcomet}, to generate error annotations. Each English source and its corresponding Spanish MT is passed through the trained QE model, which produces error spans along with associated confidence scores and severity levels (minor, major, or critical). We display the highlighted error spans with a color-coded legend (\colorbox{yellow}{  } Minor | \colorbox{orange!60}{  } Major | \colorbox{red!50}{  } Critical). Confidence scores are not shown to participants. When the identified error span is a subword segment, we highlight at the word level to improve readability. Error annotations are presented only on the MT output, reflecting how the QE model naturally operates. If the QE model does not produce any error annotations, no highlights are shown, and the following message is displayed: ``\textit{AI did not detect any errors}''. On average, each example contains 1.43 annotated error spans.



\paragraph{\raisebox{-0.2em}{\includegraphics[height=1.1em]{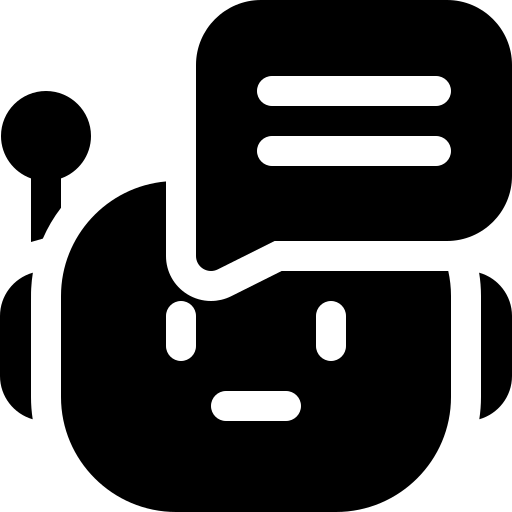}} LLM Explanation.}
We generate natural language explanations using \textsc{LLaMA-3.3 70B} \citep{llama3}.\footnote{\url{https://huggingface.co/meta-llama/Llama-3.3-70B-Instruct}} Instead of instructing the model to make a shareability decision, we prompt it to assess the overall quality of the Spanish MT relative to the English source text. The exact prompt is provided in Appendix~\ref{appendix:prompt_explanation}. For digestibility, we constrain the model to generate responses of fewer than three sentences. The generated explanations have an average length of 46.35 words.

\paragraph{\raisebox{-0.2em}{\includegraphics[height=1.1em]{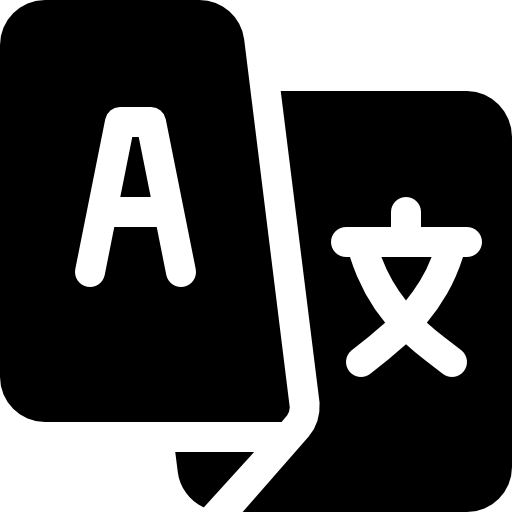}} Backtranslation.}
We use the Google Translate API\footnote{\url{https://translate.google.com/}} to backtranslate the Spanish MT output since it represents one of the most widely used consumer-facing
commercial MT systems \citep{pitman2021google}. The translation quality is reasonable, as indicated by QE scores between the Spanish MT and its backtranslation: 0.860 from \textsc{COMET-QE} \citep{rei-etal-2020-comet} and 0.962 from \textsc{xCOMET-QE XL} \citep{guerreiro-etal-2024-xcomet}. Participants are not informed about the specific MT system used; instead, they are simply shown a brief explanation stating that the backtranslation represents ``\textit{how the AI system translates the Spanish MT back into English}''.

\definecolor{qablue}{RGB}{57, 133, 219}

\paragraph{\raisebox{-0.2em}{\includegraphics[height=1.1em]{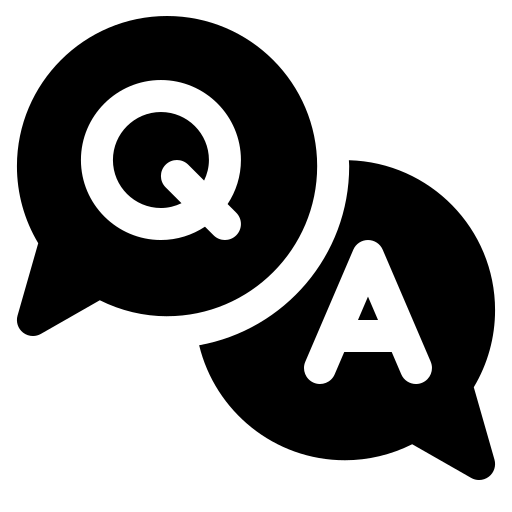}} QA Table.}
We use the \textsc{AskQE} framework \citep{ki2025askqequestionansweringautomatic} for question generation and answering, where questions are generated from the source text and answers are drawn from both the source and the backtranslated MT output. Specifically, we adopt an optimized version of \textsc{AskQE} that uses \textsc{LLaMA-3.3 70B} and entailed facts to guide question generation, and Google Translate for backtranslation of the Spanish MT output. All prompts are provided in Appendix~\ref{appendix:prompt_qa}. On average, each example yields 2.65 questions, with an average question length of 10.04 words. As illustrated in Figure~\ref{fig:feedback_types}, we present the QA pairs in a table format with the following statements: ``\textit{The questions are about the original English content}'' and ``\textit{The answers are based on two sources: the original English text and the Spanish translation, which has been translated back into English for display}''. When the two answers are identical or highly similar, they are displayed in \textcolor{orange}{orange}, else, \textcolor{qablue}{blue}. Similarity is computed using a soft variant of exact matching, with normalization for punctuation, case, whitespace, and articles.

\subsection{Stimuli Collection}
\label{sec:stimuli_collection}
We sample 40 English-Spanish examples from the \textsc{ContraTICO} dataset \citep{ki2025askqequestionansweringautomatic},\footnote{Criteria for data selection are detailed in Appendix~\ref{appendix:data_selection}.} which contains contrastive, synthetic MT errors in the COVID-19 domain. Each reference translation from the \textsc{TICO-19} dataset \citep{anastasopoulos-etal-2020-tico} is perturbed using eight linguistic perturbations, categorized as either minor or critical based on the potential real-world impact of the MT error. We recruit five bilingual annotators to independently annotate each MT output for gold shareability labels. Inter-annotator agreement, measured by Fleiss' Kappa\footnote{\url{https://en.wikipedia.org/wiki/Fleiss_kappa}}, is moderate (0.449). We select 20 examples for the study based on high agreement scores determined by majority vote, with 10 examples per label. Further details are in Appendix~\ref{appendix:gold_annotation}.\footnote{English source sentences contain 471 words in total (average 23.6 words per sentence), and the Spanish translations contain 590 words in total (average 29.5 words per sentence).}


\begin{figure*}
    \centering
    \includegraphics[width=\linewidth]{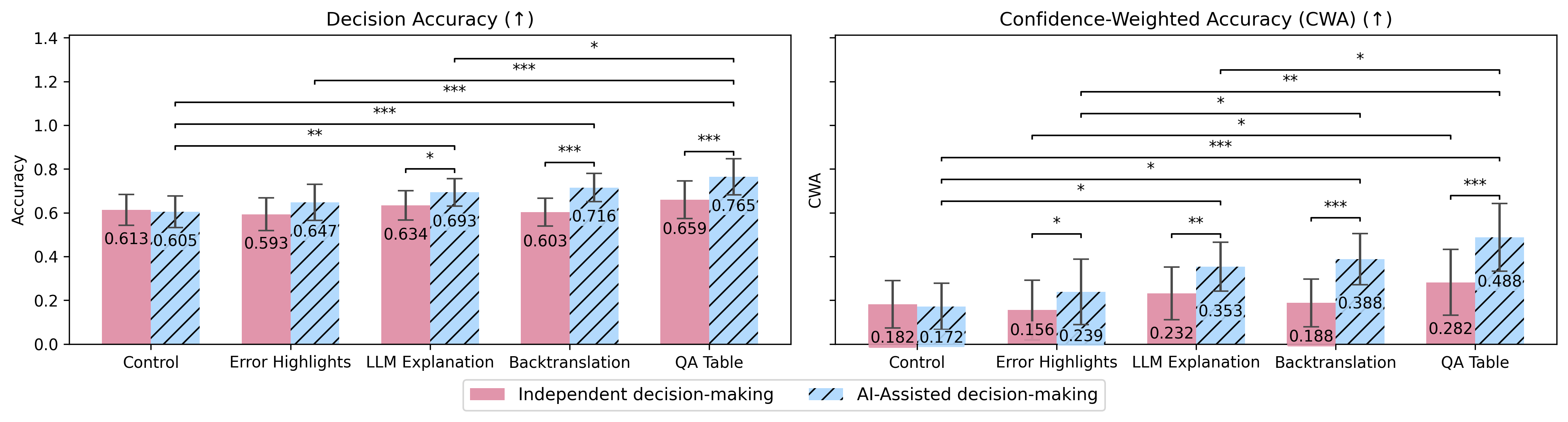}
    \caption{Average decision accuracy (\textit{left}) and \textsc{CWA} (\textit{right}) for each condition. Paired-sample $t$-tests are performed to compare independent and AI-assisted performance and linear mixed-effects ANOVA with Bonferroni corrections to compare different treatment conditions. *: significant with $p$-value < 0.05; **: $p$ < 0.01; ***: $p$ < 0.001; Non-marked: not statistically significant. Detailed results are provided in Appendix~\ref{appendix:independent_aiassisted}.}
    \label{fig:rq1}
\end{figure*}

\subsection{Participants}
\label{sec:demographics}
We recruited 91 participants residing in the United States who self-identified English as their first, primary, and fluent language. Recruitment was conducted in two stages to exclude participants proficient in Spanish: \textbf{(1)} A pre-screening survey, where participants reported their English and Spanish proficiency on a five-point scale; and \textbf{(2)} The main task, limited to those who reported high English and low Spanish proficiency in the pre-screening phase. Each main task participant received 5 USD for completing the task (equivalent to 20 USD/hour),\footnote{The task took a median of 14 minutes to complete.} and 30 participants who achieved over 70\% overall decision accuracy received an additional 2 USD bonus. Our institution’s IRB approved to conduct the study. Participants provided informed consent prior to the study. Further details are in Appendix~\ref{appendix:recruitment_process}.

Of the 91 participants, 90 reported English as their first language, and one reported both Filipino and English. The average self-reported English proficiency was 5/5 and Spanish proficiency was 1.83/5.\footnote{We detail language proficiency scale in Appendix~\ref{appendix:design_details}.} Reported monthly MT usage varied: 5 participants (5.49\%) never used MT, 24 (26.4\%) rarely used it, 32 (35.2\%) used it sometimes, 19 (20.9\%) often, and 11 (12.1\%) used MT almost every day. Data from one participant who failed both shareability checks was excluded from analysis. Participants were randomly assigned to one of five conditions, with 18 in each group.

\begin{figure*}
    \centering
    \includegraphics[width=\linewidth]{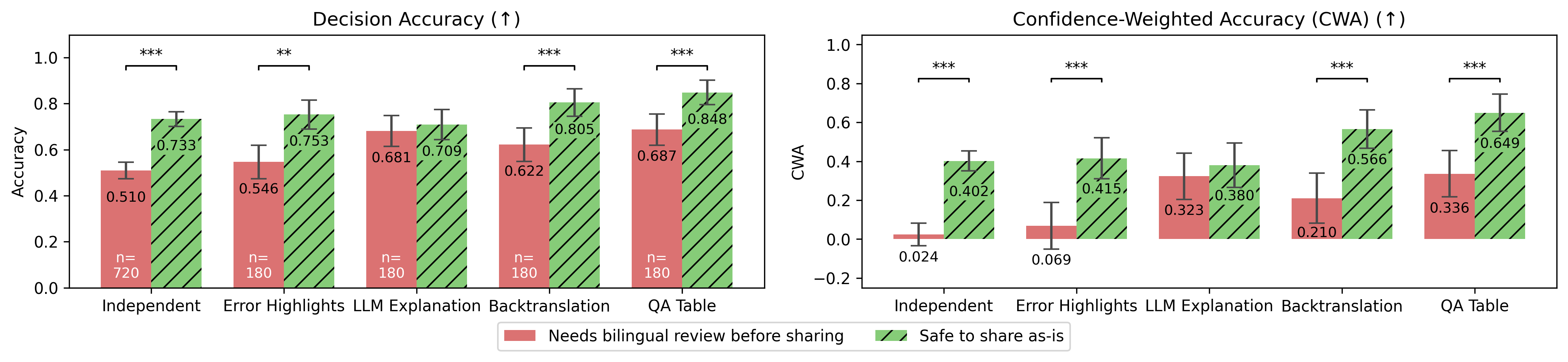}
    \caption{Average decision accuracy (\textit{left}) and \textsc{CWA} (\textit{right}) for each type of quality feedback and shareability label. $\mathbf{n}$ indicates the number of examples aggregated for each condition and label. \textbf{Independent} aggregates responses made without quality feedback across all conditions. **: statistically significant with $p$-value < 0.01; ***: $p$ < 0.001; Non-marked ones are not statistically significant. Detailed results are provided in Appendix~\ref{appendix:shareability_label}.}
    \label{fig:per_shareability}
\end{figure*}

\subsection{Dependent Variables}
\label{sec:evaluation}

\paragraph{Decision Accuracy.}
For each example $e$, we ask participants to decide whether the translation is of sufficient quality to safely share using a binary scale. We compute decision accuracy by comparing each participant's shareability judgment $\hat{s}$ against the gold label $s^*$ for each example $e \in E$:
\begin{equation}
    \mathbf{Decision Acc.}(E) = \frac{1}{|E|} \sum_{e \in E}\mathbbm{1}(\hat{s} = s^*)
\end{equation}

\paragraph{Confidence-Weighted Accuracy (CWA).}
Following \citet{mehandru-etal-2023-physician}, we combine decision accuracy and confidence scores using confidence weighting \citep{ebel1965confidence, marshall2017confidence} to evaluate whether participants made the correct decision weighted by their confidence in that decision. This metric serves as a measure of (in)appropriate reliance, where higher scores indicate accurate decisions made with well-calibrated confidence. Formally, for each example $e \in E$, we combine shareability $\hat{s}$ and confidence $c$ as follows:
\begin{equation}
\begin{aligned}
    \mathbf{CWA}(E) = \frac{1}{|E|} \sum_{e \in E} \mathrm{sign}(\hat{s}) \cdot \frac{c}{5} \\
    \mathrm{sign}(\hat{s}) =
    \begin{cases}
    1, & \text{if } \hat{s} = s^* \\
    -1, & \text{otherwise}
    \end{cases}
\end{aligned}
\end{equation}

\paragraph{Switch Percentage.}


Switch percentage is a widely used behavioral measure of reliance, capturing how often participants change their decisions after viewing AI feedback \citep{improving_awareness, stated_reliance}. In our context, it reflects how quality feedback influences final shareability judgments \citep{eckhardt2024surveyaireliance}. 
We compute three metrics following the framework of \citet{schemmer2023appropriate}: \textbf{(1) Over-reliance}: the proportion of cases where a participant changes from a correct to an incorrect decision after feedback; \textbf{(2) Under-reliance}: the proportion of cases where a participant does not change from an incorrect decision to a correct one after the quality feedback; \textbf{(3) Appropriate reliance}: the proportion of cases where a participant either corrects an incorrect decision after receiving feedback (switch) or maintains a correct decision (no switch). 
To account for the differing consequences of reliance depending on shareability, we further break down each reliance metric by shareability label, as detailed in Appendix~\ref{appendix:detailed_switch}.

\section{Results}
\label{sec:results}

We begin by comparing independent and AI-assisted decision-making performance (\S\ref{sec:results_rq1}). We then evaluate the four quality feedback types in detail, in terms of decision accuracy, \textsc{CWA} (\S\ref{sec:results_rq2}), and switch percentage (\S\ref{sec:results_rq3}).


\subsection{RQ1: Does Quality Feedback Improve MT Decision-Making?}
\label{sec:results_rq1}

We perform paired-sample $t$-tests to compare independent and AI-assisted performance. As shown in Figure~\ref{fig:rq1}, participants in all four treatment conditions generally exhibit higher decision accuracy (\textit{left}) and appropriate reliance, measured by \textsc{CWA} (\textit{right}) in the AI-assisted decision-making step compared to the independent step. We observe statistically significant gains in average decision accuracy for LLM explanation (9.31\%; \textit{p} < 0.05), backtranslation (18.7\%; \textit{p} < 0.001), and QA table feedback (16.1\%; \textit{p} < 0.001), resulting in an overall average improvement of 8.32\% across all conditions. \textsc{CWA} improves across all conditions with greater extent, averaging 15.3\%, indicating that providing any quality feedback is more effective at helping participants make accurate decisions with well-calibrated confidence than at improving decision accuracy alone.

We further find that both decision accuracy (M = 0.605, S.E. = 0.036) and \textsc{CWA} (M = 0.172, S.E. = 0.054) in the AI-assisted step are significantly lower in the control condition than in all treatment conditions except for the error highlights group. Moreover, the within-group difference between the independent and AI-assisted steps in the control condition is not statistically significant. This indicates that our two-step setup does not induce learning effects but the observed gains in decision accuracy and appropriate reliance stem from the quality feedback, not from repetition.

\subsection{RQ2: Which Feedback is Most Effective?}
\label{sec:results_rq2}

We perform linear mixed-effects \textsc{ANOVA}, followed by Bonferroni correction for multiple comparisons across treatment conditions. Implicit feedback types generally outperform explicit ones. Participants who received QA table feedback have significantly higher AI-assisted decision accuracy (M = 0.765, S.E. = 0.022) than those in the error highlights (M = 0.647, S.E. = 0.043; \textit{p} < 0.01) and the LLM explanation group (M = 0.693, S.E. = 0.042; \textit{p} < 0.05), as shown in Figure~\ref{fig:rq1}. No significant difference was found between QA table and backtranslation (M = 0.716, S.E. = 0.025).

A similar pattern emerges for appropriate reliance (\textsc{CWA}). QA table group achieved significantly higher \textsc{CWA} (M = 0.488, S.E. = 0.040) than error highlights (M = 0.239, S.E. = 0.071; \textit{p} < 0.01) and LLM explanations (M = 0.353, S.E. = 0.070; \textit{p} < 0.05). Backtranslation (M = 0.388, S.E. = 0.072; \textit{p} < 0.05) also yielded significantly higher \textsc{CWA} than error highlights. These results suggest that QA table feedback is the most effective overall, outperforming both explicit feedback types in supporting accurate and well-calibrated MT decisions.

Participants' independent decision-making performance did not significantly differ across conditions, except for \textsc{CWA}, where QA table feedback (M = 0.282, S.E. = 0.040) significantly outperform error highlights (M = 0.156, S.E. = 0.041; \textit{p} < 0.05).

Overall, our findings show that all three quality feedback types except error highlights significantly improve both decision accuracy and appropriate reliance compared to no feedback, with implicit feedback types (backtranslation and QA table) showing stronger and more consistent statistical effects (\S\ref{sec:results_rq1}). Among them, QA table consistently yields the greatest gains in both metrics (\S\ref{sec:results_rq2}).



\subsection{RQ3: Which Feedback do Users Rely on Most Appropriately?}
\label{sec:results_rq3}

\begin{figure}
    \centering
    \includegraphics[width=\linewidth]{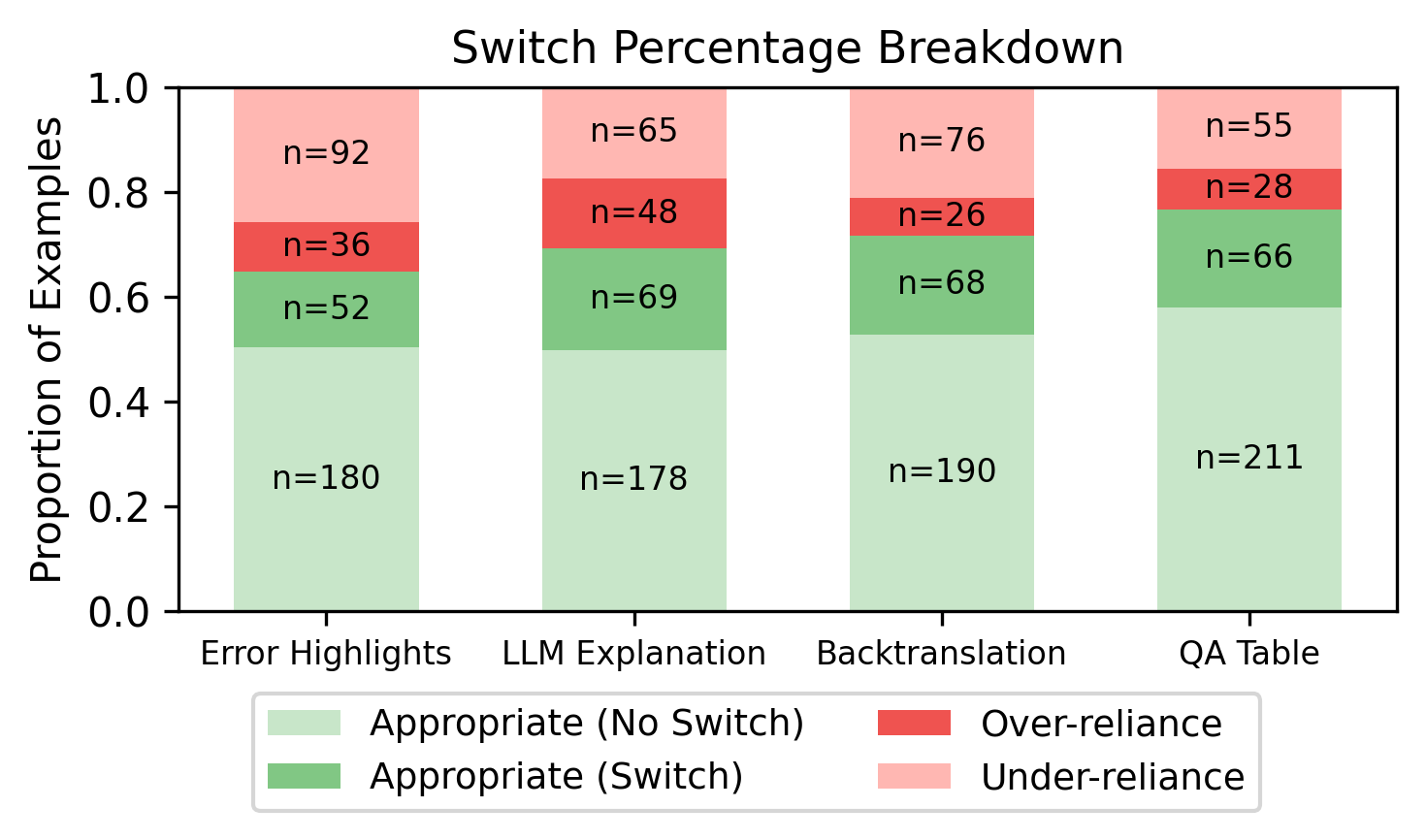}
    \caption{Breakdown of switch percentages by quality feedback type, showing appropriate, over-, and under-reliance.}
    \label{fig:switch_percent}
\end{figure}

For each quality feedback, we compute switch percentage to capture participants' behavioral patterns of reliance in Figure~\ref{fig:switch_percent}. Under-reliance is highest in the error highlights group (25.6\%), followed by backtranslation (21.1\%), LLM explanation (18.1\%), and QA table (15.3\%). Interestingly, implicit feedback types (QA table (7.78\%) and backtranslation (7.22\%)) yield lower over-reliance than explicit ones (LLM explanation (13.3\%) and error highlights (10.0\%)). Across all conditions, participants are more likely to maintain their initial decisions (regardless of correctness) than to change them, as under-reliance consistently exceeds over-reliance, and appropriate reliance (no switch) exceeds (switch).

We further examine switch percentages by shareability label in Appendix~\ref{appendix:detailed_switch}. Participants show higher over-reliance and lower under-reliance for shareable examples than non-shareable ones.\footnote{We refer to \textit{shareable} as examples labeled ``Safe to share as-is,'' and \textit{non-shareable} as those labeled ``Needs bilingual review before sharing''.} This suggests that participants are more likely to change their decisions when initially judging translations as shareable, but tend to maintain their decisions when judging them as non-shareable.





\section{Analysis}
\label{sec:analysis}


\subsection{Shareable vs. Non-shareable MT}
\label{sec:shareability}

As shown in Figure~\ref{fig:per_shareability}, treatment condition participants consistently achieve significantly higher decision accuracy (\textit{p} < 0.01 for error highlights; \textit{p} < 0.001 for others) and \textsc{CWA} scores (\textit{p} < 0.001) on shareable examples than on non-shareable ones. One exception is the LLM explanation group, where the differences are statistically not significant. This suggests that participants generally make more accurate and appropriate decisions when evaluating good translations than problematic ones, indicating that helping users reliably identify critical MT errors remains a challenge.

\begin{table}
\centering
\resizebox{\linewidth}{!}{%
    \begin{tabular}{llllllll}
    \specialrule{1.3pt}{0pt}{0pt}
    \textbf{Condition} & \textbf{Mental burden} & \textbf{Helpfulness} & \textbf{Trust} \\
    \toprule

    \textbf{Control} & 5.83 & 2.89 & - \\

    \textbf{Error Highlights} & 4.94 & 3.83 & 3.89 \\
    
    \textbf{Explanation} & 4.06 & \textbf{4.39} & 4.11 \\

    \textbf{Backtranslation} & 4.06 & \textbf{4.39} & 4.06 \\

    \textbf{QA Table} & \textbf{4.00} & \textbf{4.39} & \textbf{4.22} \\

    \specialrule{1.3pt}{0pt}{0pt}
    \end{tabular}
}
\caption{Average mental burden (1-7, ↓), helpfulness (1-5, ↑), and trust for future use (1-5, ↑) for each condition group. Best scores for each metric are \textbf{bold}.}
\label{tab:post_survey}
\end{table}

\subsection{Self-reported Perception vs. Actual Performance}
\label{sec:perception}
In the post-task survey, participants rated the quality feedback they received in terms of perceived mental burden, helpfulness, and trust for future use. As shown in Table~\ref{tab:post_survey}, the control condition group reported highest mental burden (5.83) and lowest helpfulness (2.89), suggested that repeated exposure without any quality feedback increased cognitive load without enhancing perceived utility.

Among the four treatment conditions, QA table group reported the highest level of trust (4.22), aligning with findings from Section~\ref{sec:results_rq2} that this feedback is most effective at improving appropriate reliance.
In contrast, participants who received error highlights gave the lowest ratings for helpfulness (3.83) and trust (3.89), which is consistent with their relatively poor performance in both decision accuracy and appropriate reliance (\textsc{CWA}).
Interestingly, while LLM explanations do not yield large gains in decision accuracy or \textsc{CWA}, they have relatively high ratings for helpfulness (4.39) and trust (4.11), which may reflect the over-reliance discussed in Section~\ref{sec:results_rq3}. 
In terms of mental burden, the error highlights group reported higher score (4.94) than other groups. We attribute this to the nature of error highlights as a \textit{target}-side feedback mechanism \citep{leiter-towards}, which displays highlights on the Spanish MT output, making it difficult to interpret for monolingual source speakers.


\subsection{What Makes Quality Feedback \textit{Helpful}?}
\label{sec:free_comments}

We present treatment condition participants’ responses on how they used quality feedback in their decision-making and which aspects they found helpful or unhelpful.
For error highlights, two participants found them largely unhelpful, noting that the highlights were only shown on the Spanish MT, illustrating a key limitation of a target-side feedback. However, some appreciated the explicitness of the highlights, stating that they pointed to ``\textit{areas that not have been accurately translated}'' or ``\textit{key translation mistakes}''. Similarly, four participants valued LLM explanations for showing ``\textit{exactly what is correct or incorrect}''. Some further appreciated for offering insights into alternative translations or contextual relevance.

In contrast, participants who received implicit feedback described a more self-directed decision-making process. Backtranslation was considered helpful for verifying ``\textit{some tiny unsure details}'' or checking whether ``\textit{the core meaning of the original English text was preserved in the Spanish MT}'' by comparing the two English texts. Similarly, QA table encouraged participants to revisit the MT output when a mismatch was detected (``\textit{if the statement is blue, I double check the phrase again}''), ``\textit{compare the words and make determinations}'' themselves. Detailed comments are provided in Appendix \ref{appendix:detailed_comments}.

\section{Conclusion}


We explore the utility of quality feedback in helping monolingual source speakers make reliable MT decisions. We conduct a between-subjects human study where participants decide whether Spanish MT outputs are safe to share, first independently and then with one of five conditions. The four treatment conditions include different types of MT quality feedback: two explicit (error highlights and LLM explanation) and two implicit (backtranslation and QA table).

We find that all feedback types except error highlights significantly improve decision accuracy and appropriate reliance (\S\ref{sec:results_rq1}). Implicit feedback, especially QA table, outperforms explicit feedback in both objective performance and self-reported ratings (\S\ref{sec:results_rq2}, \S\ref{sec:perception}) and while explicit feedback prompts more decision changes, it also increases over-reliance (\S\ref{sec:results_rq3}). We further show that participants are better at confirming good translations than detecting problematic ones (\S\ref{sec:shareability}). 

Overall, our findings underscore the value of feedback that guides users' implicit interpretation rather than prescribing decisions. Implicit methods may be especially effective, as they preserve users' agency in the decision-making process \citep{savoldi2025translationhandsmanycenteringlay, yimin_2025_sustaining}.
Along with insights into what participants found helpful across feedback types (\S\ref{sec:free_comments}), our work calls for further research exploring feedback to help users reliably identify critical MT errors in realistic use cases.

\section{Limitations}

\paragraph{Presentation differences.}
We acknowledge that presentation differences, such as formatting or visual salience, can influence user behavior independently of the underlying feedback type. To mitigate this, we iteratively designed and piloted each feedback interface to ensure clarity and minimize usability discrepancies across conditions. Specifically, we conducted seven rounds of internal pilot testing and usability checks to identify sources of confusion, refine language, or layout. While some format-specific differences were necessary (\textit{e.g.,} error highlights inherently rely on color to convey span-level quality signals, while QA tables require side-by-side comparisons), we carefully calibrated to ensure that each condition represented a ``best of its breed'' version of that feedback type. Therefore, although we acknowledge the possibility of minor usability-related effects, we believe these differences are unlikely to fully account for the performance gaps observed between feedback types.

\paragraph{Study with monolingual target speakers.}
Our human study focuses on monolingual source speakers who self-identified as proficient in English but not in Spanish, simulating a scenario in which non-English speakers encounter MT outputs of COVID-19 articles originally written in English. An important complementary study remains \---\ evaluating with the monolingual target speakers. This would require modifying our current quality feedback setup: \textbf{(1)} Error highlights would continue to be displayed on the MT output; \textbf{(2)} LLM explanations would need to be presented in the target language; \textbf{(3)} Backtranslation would no longer be a suitable feedback; and \textbf{(4)} For QA table, questions would need to be generated from the backtranslated source, with answers derived from both the backtranslation and the (possibly perturbed) MT output. Future work can explore this variant through a human-centered study to assess how different forms of quality feedback influence MT decision-making for monolingual target speakers.

\paragraph{Limited scope.}
The scope of our study is limited to the current experimental setup. While we focus on the English-Spanish language pair, motivated by Spanish being the most widely spoken non-English language in the United States \citep{census_languages_2022}, our findings may not generalize to other language pairs. For consistency and fair comparison, we use the same model (\textsc{LLaMA-3.3 70B}) to generate both LLM explanations and QA table feedback, and the same MT system (Google Translate) to produce backtranslations and the backtranslated MT outputs used in QA generation. Our evaluation is also limited to four types of quality feedback: error highlights, LLM explanations, backtranslation, and QA table.

Moreover, we intentionally focused on a single domain (communicating COVID-19 protocols) and a specific user population with specific language background in order to ensure that our study reflects realistic decision-making scenarios. This controlled setting helped participants better operationalize the notion of shareability and limited potential confounding factors in the study. We believe the core findings ought to generalize to other language pairs since MT quality was not a major factor, and possibly other domains with similar notions of shareability. 

Understanding the effect of quality feedback on MT reliance across a wider range of contexts is an important direction of future work. For instance, the doctor-patient setting represents a valuable use case where the doctor as a user of MT has more consideration and technical expertise in deciding whether MT outputs are shareable. Another example could be dialogues, where additional rounds of human and automatic feedback is possible. We view our study as a foundational step toward more extensive future research in this area.

\paragraph{Use of synthetic dataset.}
While our study aimed to simulate a realistic decision-making scenario, the MT outputs themselves are drawn from a synthetically constructed dataset \textsc{ContraTICO}. This design choice allowed us to systematically control for specific error types and severity levels. However, we acknowledge that these errors may not fully capture the complexity and variability of naturally occurring MT errors.

\section*{Ethics Statement}

This study was approved by our university’s Institutional Review Board (IRB). All participants provided informed consent prior to participation and were compensated according to the rates specified in the consent form. Additionally, participants who achieved an overall accuracy above 70\% received a performance-based incentive.
\section*{Acknowledgements}

We would like to thank the members of the \textsc{clip} lab at University of Maryland for their constructive feedback and support.
Specifically, we are grateful to Hiba El Oirghi, Yu Hou, HyoJung Han, Kartik Ravisankar, Nishant Balepur, Navita Goyal, and Calvin Bao for their help in testing the user interface. We especially appreciate Yimin Xiao for extensive discussions on the statistical analysis in the paper. This work was supported in part by the Human Language Technology Center of Excellence at Johns Hopkins University, by NSF Fairness in AI Grant 2147292, and by the Institute for Trustworthy AI in Law and Society (TRAILS), which is supported by the National Science Foundation under Award No. 2229885. The views and conclusions contained herein are those of the authors and should not be interpreted as necessarily representing the official policies, either expressed or implied, of NSF or the U.S. Government. The U.S. Government is authorized to reproduce and distribute reprints for governmental purposes notwithstanding any copyright annotation therein.

\bibliography{custom}

\clearpage
\appendix
\section{Prompts}
\subsection{LLM Explanation}
\label{appendix:prompt_explanation}
We show prompt used for generating explanations with \textsc{LLaMA-3.3 70B} \citep{llama3}.

\begin{prompt}[title={Prompt A.1: LLM Explanation}]
\textbf{Task:} Your task is to evaluate the quality of the Spanish translation of the English sentence. Give your explanation in less than 3 sentences. \\ \\
\textbf{English sentence:} \texttt{\{source\}} \\
\textbf{Spanish translation:} \texttt{\{target\}} \\
\textbf{Explanation:} 
\end{prompt}

\subsection{Question-Answer (QA) Table}
\label{appendix:prompt_qa}
We use the same prompts for both question generation (A.2.1) and answering (A.2.2) as those used in \textsc{AskQE} \citep{ki2025askqequestionansweringautomatic}.

\begin{prompt}[title={Prompt A.2.1: Question Generation (QG)}]
\textbf{Task:} You will be given an English sentence and a list of atomic facts, which are short sentences conveying one piece of information. Your goal is to generate a list of relevant questions based on the sentence. Output the list of questions in Python list format without giving any additional explanation. \\ \\
*** Example Starts *** \\
\textbf{Sentence:} It was declared a pandemic by the World Health Organization (WHO) on 11 March 2020. \\
\textbf{Atomic facts:} [``It was declared a pandemic on 11 March 2020.'', ``The World Health Organization (WHO) declared it a pandemic.'] \\
\textbf{Questions:} [``What was declared on 11 March 2020?", "Who declared it a pandemic?''] \\ \\
\textbf{Sentence:} The number of accessory proteins and their function is unique depending on the specific coronavirus. \\
\textbf{Atomic facts:} [``The number of accessory proteins is unique depending on the specific coronavirus.'', ``The function of accessory proteins is unique depending on the specific coronavirus.''] \\
\textbf{Questions:} [``What is unique depending on the specific coronavirus?'', ``What is unique about the function of accessory proteins?''] \\
*** Example Ends *** \\ \\
\textbf{Sentence:} \texttt{\{sentence\}} \\
\textbf{Atomic facts:} \texttt{\{atomic facts\}} \\
\textbf{Questions:}
\end{prompt}
\begin{prompt}[title={Prompt A.2.2: Question Answering (QA)}]
\textbf{Task:} You will be given an English sentence and a list of relevant questions. Your goal is to generate a list of answers to the questions based on the sentence. Output only the list of answers in Python list format without giving any additional explanation. \\ \\
*** Example Starts *** \\
\textbf{Sentence:} Some patients have very mild symptoms, similar to a cold. \\
\textbf{Questions:} [``What kind of symptoms do some patients have?'', ``What are the symptoms similar to?''] \\
\textbf{Answers:} [``Very mild symptoms'', ``A cold''] \\ \\
\textbf{Sentence:} Diabetes mellitus (784, 10.9\%), chronic lung disease (656, 9.2\%), and cardiovascular disease (647, 9.0\%) were the most frequently reported conditions among all cases. \\
\textbf{Questions:} [``What were the most frequently reported conditions among all cases?'', ``What percentage of cases reported diabetes mellitus?'', ``What percentage of cases reported chronic lung disease?'', ``What percentage of cases reported cardiovascular disease?''] \\
\textbf{Answers:} [``Diabetes mellitus, chronic lung disease, and cardiovascular disease'', ``10.9\%'', ``9.2\%'', ``9.0\%''] \\
*** Example Ends *** \\ \\
\textbf{Sentence:} \texttt{\{sentence\}} \\
\textbf{Questions:} \texttt{\{questions\}} \\
\textbf{Answers:} 
\end{prompt}

\section{Dataset Details}
\subsection{Gold Annotation}
\label{appendix:gold_annotation}
\begin{figure*}
    \centering
    \setcounter{subfigure}{0}
    \subfigure[Task Instructions]{{\includegraphics[width=0.8\textwidth]{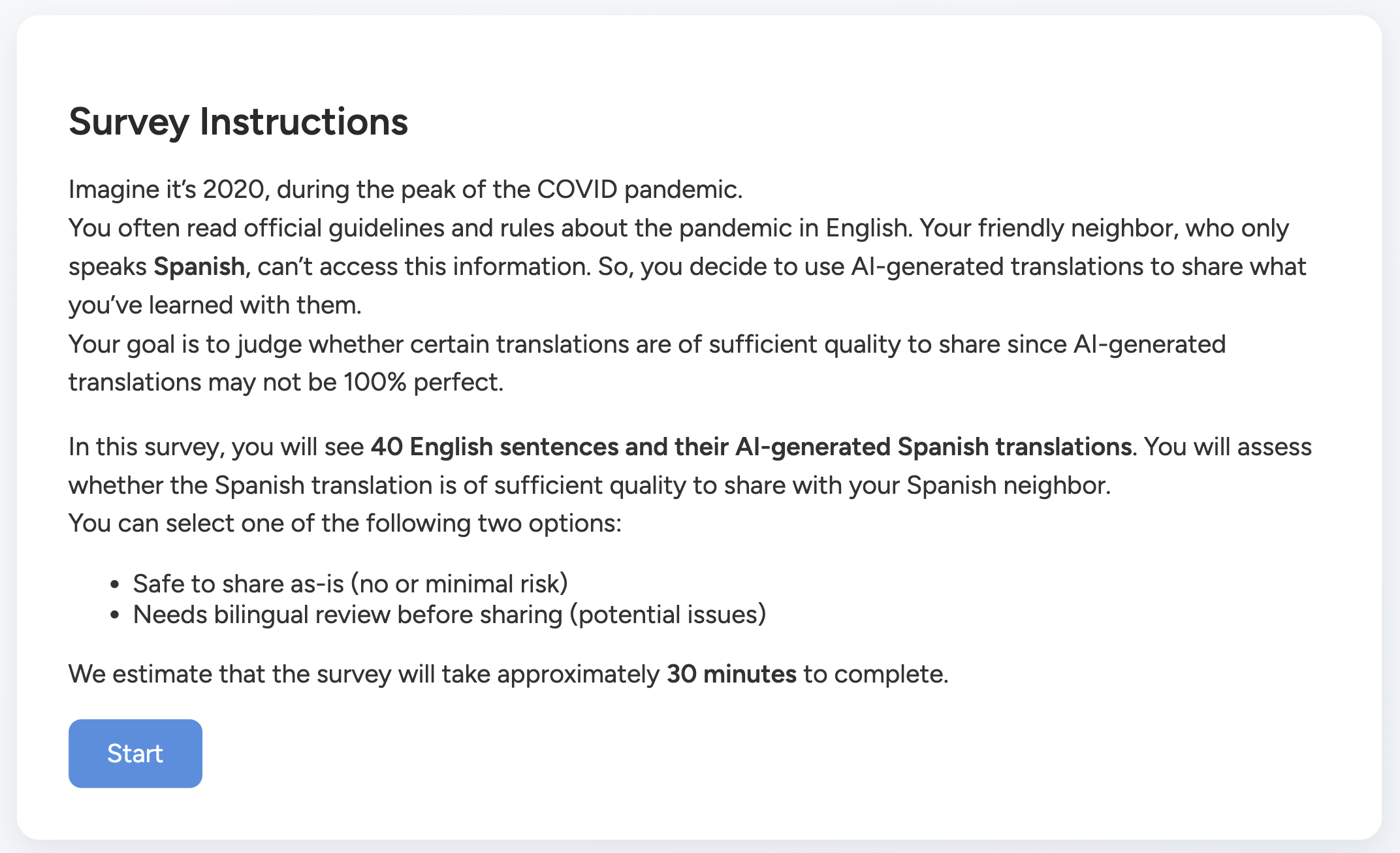}}}
    \subfigure[Example Question]{{\includegraphics[width=0.6\textwidth]{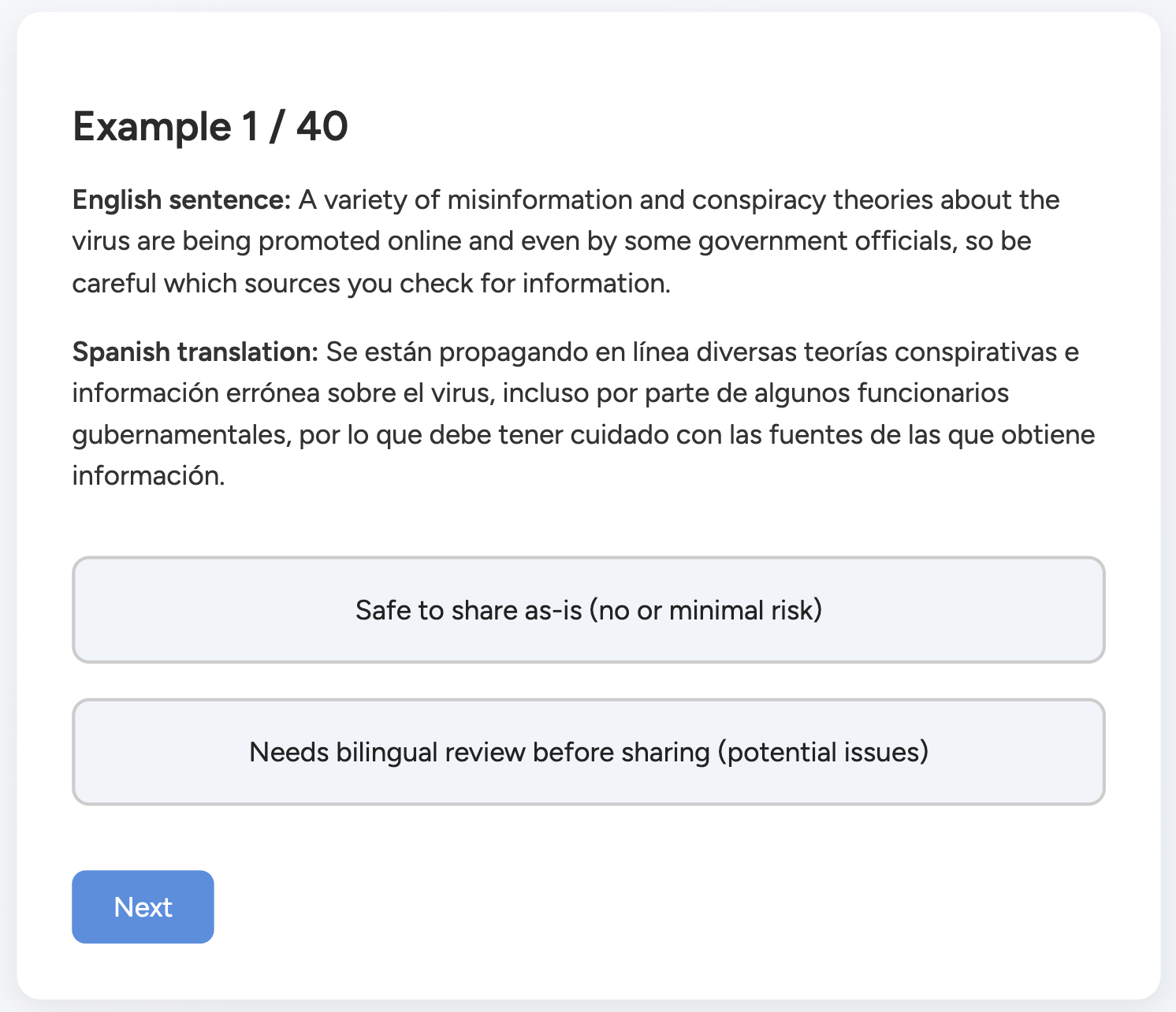}}} 

    \caption{Screenshots of the instructions provided to bilingual annotators, along with an example question.}
    \label{fig:gold_annotation}
\end{figure*}

We provide details of the gold annotation process used to collect gold shareability labels. We use Qualtrics\footnote{\url{https://www.qualtrics.com}} to design the survey and Prolific\footnote{\url{https://www.prolific.com}} to recruit annotators fluent in both English and Spanish. A total of 40 examples, each consisting of an English sentence and its Spanish translation, were presented in randomized order. As shown in Figure~\ref{fig:gold_annotation}, annotators were asked to judge whether the Spanish translation was \textbf{(1)} Safe to share as-is or \textbf{(2)} Needs bilingual review before sharing. The survey took a median completion time of 30 minutes. We recruited 5 annotators and compensated each with 8 USD (equivalent to 16 USD/hour).

We select 20 examples with high agreement scores (based on majority vote) for use in the main task. The final set includes 10 examples per label. The average agreement score is 1.0 for examples labeled as ``Safe to share as-is'' (shareable) and 0.8 for those labeled as ``Needs bilingual review before sharing'' (non-shareable). The shareable set includes 5 non-error examples and 5 minor error examples, comprising 2 synonym, 2 word order, and 1 spelling error, based on the error taxonomy in \citet{ki2025askqequestionansweringautomatic}. The non-shareable set includes 1 minor error (intensifier) and 9 critical errors: 6 alteration, 2 omission, and 1 expansion with impact error examples.

\subsection{Data Selection Criteria}
\label{appendix:data_selection}
We ensure the selected examples from the \textsc{ContraTICO} dataset \citep{ki2025askqequestionansweringautomatic} meet the following criteria before running the gold annotation: \textbf{(1)} Examples have a balanced distribution across three error severity levels: no error, minor errors, and critical errors in MT; \textbf{(2)} Examples have comparable lengths across error severity levels to minimize the influence of sentence length on participants' confidence \citep{zouhar-bojar-2020-outbound}; \textbf{(3)} Examples are relevant to the scenario context. We focus on the Wikivoyage\footnote{\url{https://www.wikivoyage.org/}} subset of the dataset, which contains announcements and protocols related to COVID-19 (e.g., ``It was declared a pandemic by the World Health Organization (WHO) on 11 March 2020.'') instead of informal conversations (e.g., ``and does this pain move from your chest?'') or sentences with highly technical terms (e.g., ``Like all coronaviruses, virions consist of single-stranded positive-sense RNA enclosed within an envelope.'').

\section{Study Design Details}
\label{appendix:design_details}
We present details about our human study design.
\subsection{Annotation Interface}
\label{appendix:annotation_interface}
\begin{figure*}
    \centering
    \setcounter{subfigure}{0}
    \subfigure[Pre-task Survey]{\includegraphics[width=0.49\textwidth]{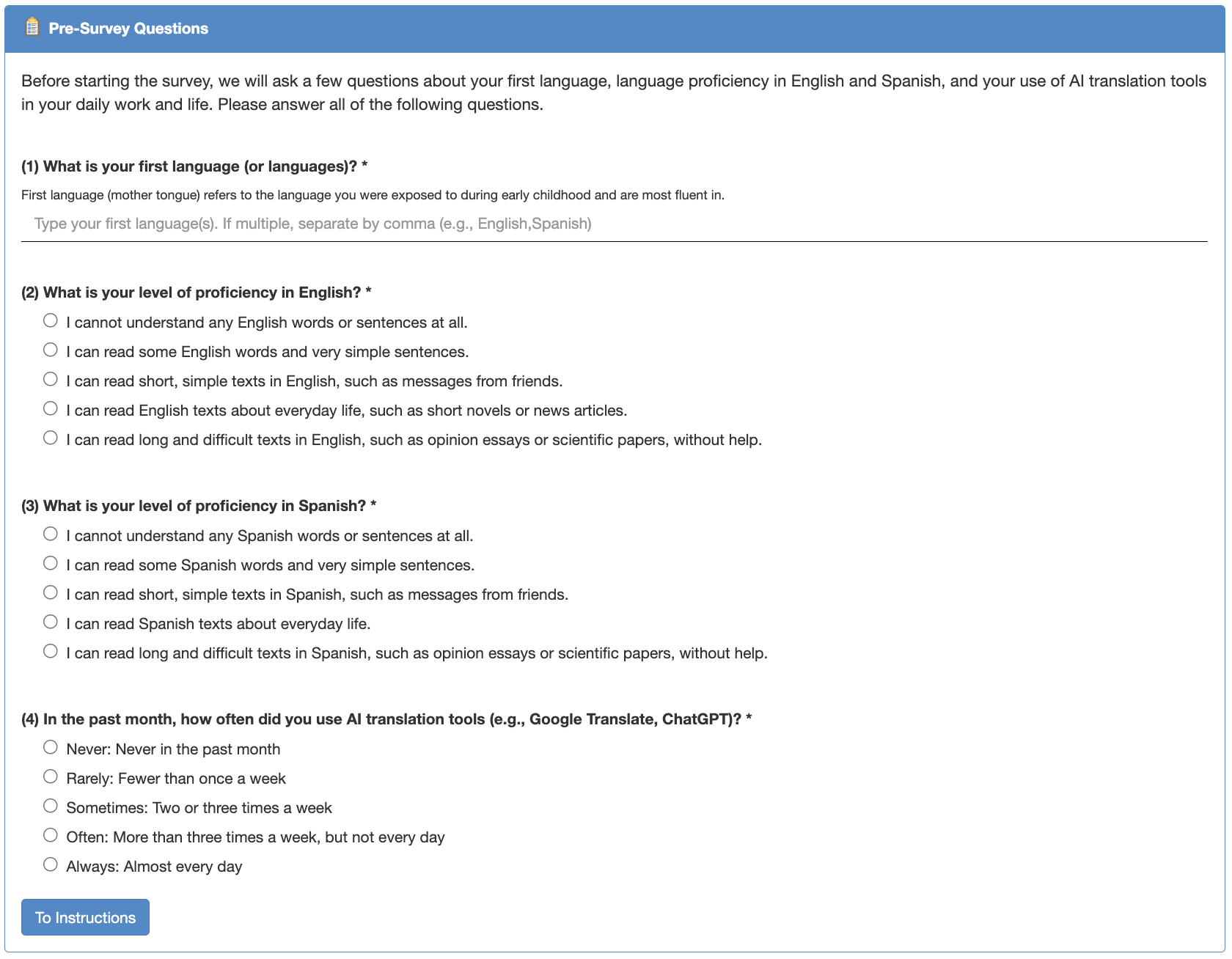}}
    \subfigure[Task Instructions]{\includegraphics[width=0.49\textwidth]{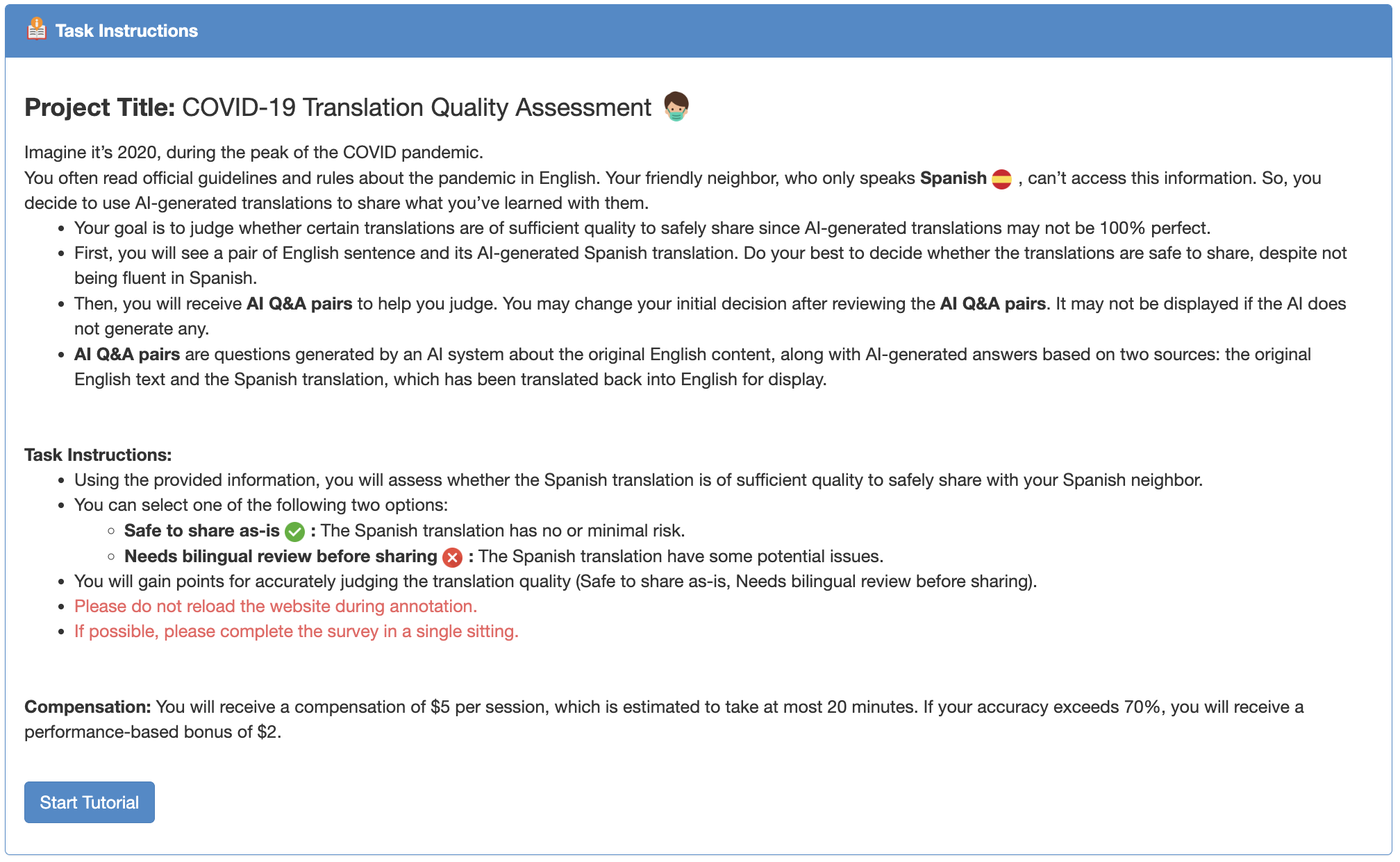}} 
    
    \subfigure[Tutorial]{\includegraphics[width=0.49\textwidth]{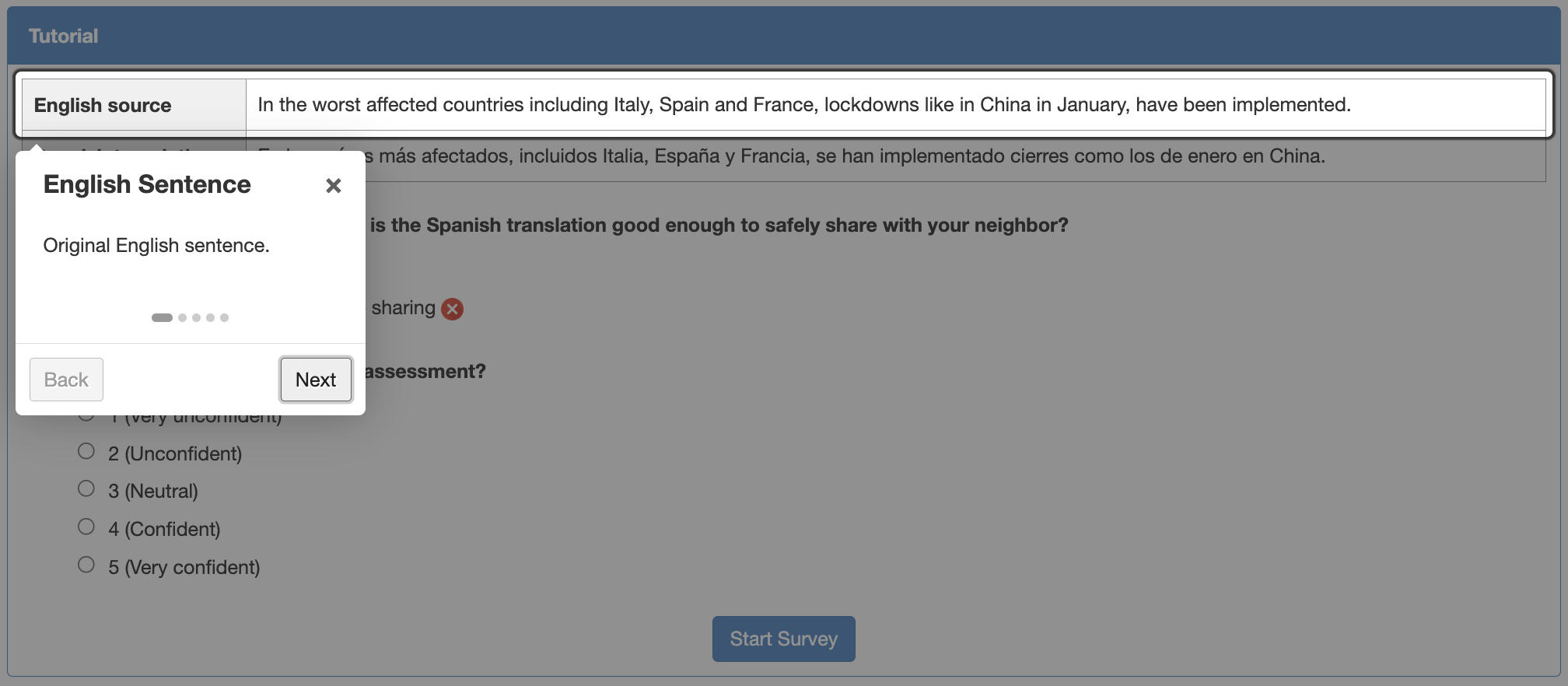}}
    \subfigure[Independent Decision-Making]{\includegraphics[width=0.49\textwidth]{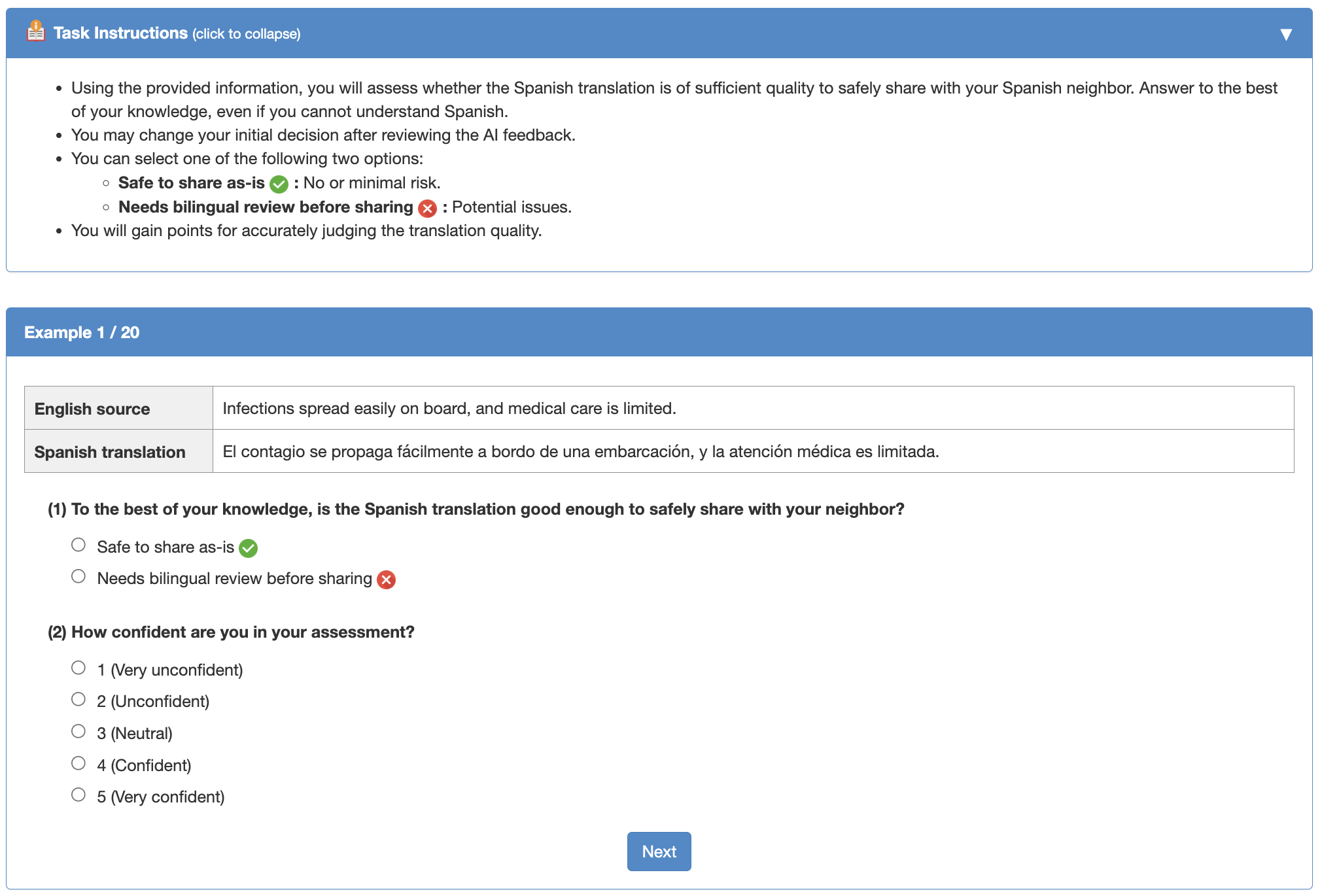}} 
    
    \subfigure[AI-Assisted: \raisebox{-0.2em}{\includegraphics[height=1.1em]{figures/logos/annotation.png}} Error Highlights]{\includegraphics[width=0.49\textwidth]{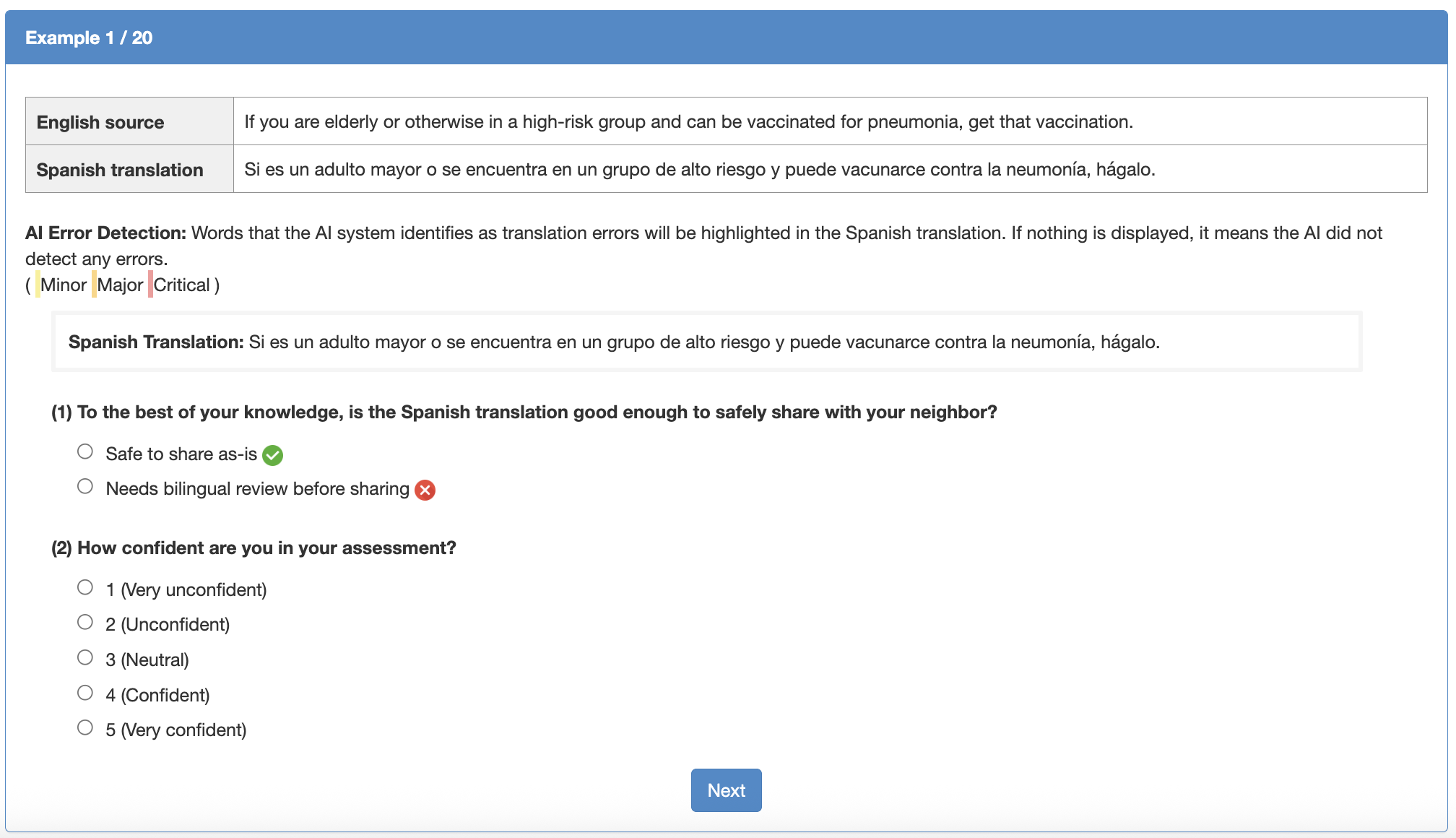}} 
    \subfigure[AI-Assisted: \raisebox{-0.2em}{\includegraphics[height=1.1em]{figures/logos/explanation.png}} LLM Explanation]{\includegraphics[width=0.49\textwidth]{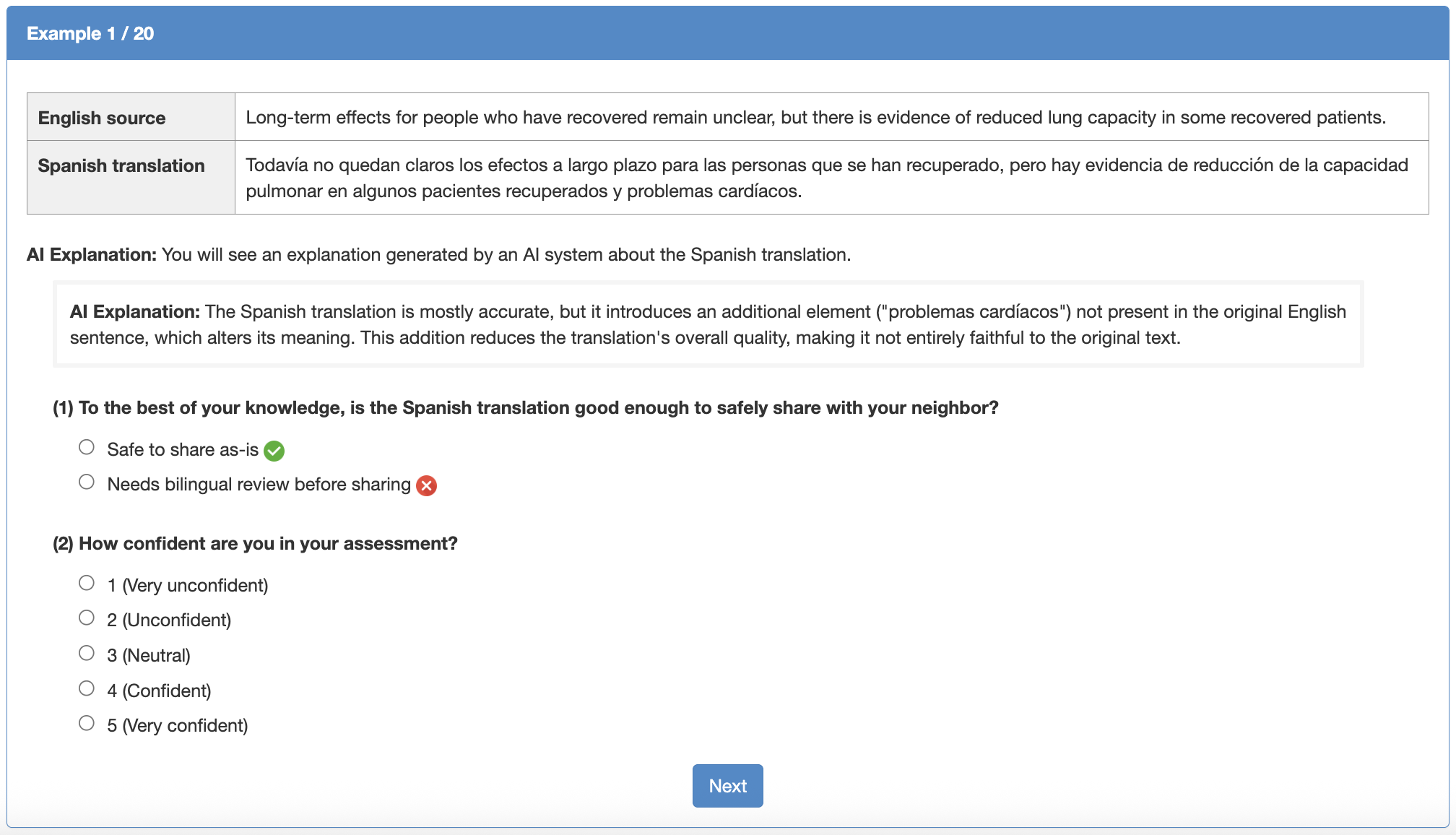}}

    \caption{Screenshots of our annotation interface, organized according to the task flow. Each example has a brief summary of the task instructions at the top, which participants can click to expand or collapse.}
    \label{fig:interface}
\end{figure*}

We built a custom annotation interface, with screenshots shown in Figure~\ref{fig:interface} following the task flow: \textbf{(1)} Consent to Participate, \textbf{(2)} Pre-task survey, \textbf{(3)} Task instructions and compensation details, \textbf{(4)} Tutorial, \textbf{(5)} Independent decision-making task, \textbf{(6)} AI-assisted decision-making task, and \textbf{(7)} Post-task survey. 
Participants were required to answer all pre- and post-task survey questions. We present an interactive tutorial before the main task to ensure that participants understand the questions they will be asked for each example. We also illustrate how each type of quality feedback was presented during the AI-assisted decision-making step.

\paragraph{Pre-Task Survey.} The pre-task survey includes the following four questions:
\begin{itemize}[leftmargin=*, itemsep=2pt, parsep=-1pt]
 \item \textbf{First language:} What is your first language (or languages)?
 \item \textbf{Proficiency in English:} What is your level of proficiency in English?
 \vspace{3pt}
 {\begin{itemize}[label=$\circ$, leftmargin=2em, itemsep=2pt, parsep=0pt]
    \small
    \item I cannot understand any English words or sentences at all.
    \item I can read some English words and very simple sentences.
    \item I can read short, simple texts in English, such as messages from friends.
    \item I can read English texts about everyday life, such as short novels or news articles.
    \item I can read long and difficult texts in English, such as opinion essays or scientific papers, without help.
  \end{itemize}}
 \item \textbf{Proficiency in Spanish:} What is your level of proficiency in Spanish?
 \vspace{3pt}
 {\begin{itemize}[label=$\circ$, leftmargin=2em, itemsep=2pt, parsep=0pt]
    \small
    \item I cannot understand any English words or sentences at all.
    \item I can read some English words and very simple sentences.
    \item I can read short, simple texts in English, such as messages from friends.
    \item I can read English texts about everyday life, such as short novels or news articles.
    \item I can read long and difficult texts in English, such as opinion essays or scientific papers, without help.
  \end{itemize}}
  \item \textbf{AI Translation Tool Usage:} In the past month, how often did you use AI translation tools (e.g., Google Translate, \textsc{ChatGPT})?
  \vspace{3pt}
  {\begin{itemize}[label=$\circ$, leftmargin=2em, itemsep=2pt, parsep=0pt]
    \small
    \item Never: Never in the past month
    \item Rarely: Fewer than once a week
    \item Sometimes: Two or three times a week
    \item Often: More than three times a week, but not every day
    \item Always: Almost every day
  \end{itemize}}
\end{itemize}

\paragraph{Post-Task Survey (Control).} In the post-task survey for the control condition, we include the following two questions:
\begin{itemize}[leftmargin=*, itemsep=2pt, parsep=-1pt]
 \item \textbf{Mental Workload:} How much mental burden (e.g., thinking, deciding) did you experience while doing the task?
 \vspace{3pt}
 {\begin{itemize}[label=$\circ$, leftmargin=2em, itemsep=2pt, parsep=0pt]
    \small
    \item 1 (Low)
    \item ...
    \item 7 (High)
  \end{itemize}}
 \item \textbf{Helpfulness:} How helpful was seeing the same example twice in your Spanish translation quality assessment?
 \vspace{3pt}
 {\begin{itemize}[label=$\circ$, leftmargin=2em, itemsep=2pt, parsep=0pt]
    \small
    \item 1 (Very unhelpful)
    \item 2 (Unhelpful)
    \item 3 (Neutral)
    \item 4 (Helpful)
    \item 5 (Very helpful)
  \end{itemize}}
\end{itemize}

\paragraph{Post-Task Survey (Treatment).} In the post-task survey for treatment conditions, we dynamically replace \underline{\textsc{feedback}} with the participant's assigned quality feedback type. The survey includes the following three questions:
\begin{itemize}[leftmargin=*, itemsep=2pt, parsep=-1pt]
 \item \textbf{Mental Workload:} How much mental burden (e.g., thinking, deciding) did you experience while using the \underline{\textsc{feedback}}?
 \vspace{3pt}
 {\begin{itemize}[label=$\circ$, leftmargin=2em, itemsep=2pt, parsep=0pt]
    \small
    \item 1 (Low)
    \item ...
    \item 7 (High)
  \end{itemize}}
 \item \textbf{Helpfulness:} How helpful was the \underline{\textsc{feedback}} in assisting the Spanish translation quality assessment?
 \vspace{3pt}
 {\begin{itemize}[label=$\circ$, leftmargin=2em, itemsep=2pt, parsep=0pt]
    \small
    \item 1 (Very unhelpful)
    \item 2 (Unhelpful)
    \item 3 (Neutral)
    \item 4 (Helpful)
    \item 5 (Very helpful)
  \end{itemize}}
 \item \textbf{Trust for Future Use:} Would you use the \underline{\textsc{feedback}} again in the future?
 \vspace{3pt}
 {\begin{itemize}[label=$\circ$, leftmargin=2em, itemsep=2pt, parsep=0pt]
    \small
    \item 1 (Very unlikely)
    \item 2 (Unlikely)
    \item 3 (Neutral)
    \item 4 (Likely)
    \item 5 (Very likely)
  \end{itemize}}
\end{itemize}
Depending on the participant's response to the \textbf{Helpfulness} question, we present an optional follow-up question:
\begin{itemize}[leftmargin=*, itemsep=2pt, parsep=-1pt]
 \item \textbf{If the response is 1 or 2:} In what ways was the information provided by the \underline{\textsc{feedback}} confusing or unhelpful?
 \item \textbf{If the response is 3:} How did you use the information provided by the \underline{\textsc{feedback}} in your assessment?
 \item \textbf{If the response is 4 or 5:} In what ways was the information provided by the \underline{\textsc{feedback}} helpful?
\end{itemize}

\subsection{Recruitment Process}
\label{appendix:recruitment_process}

\paragraph{Pre-screening Survey.}
We conducted our user studies on the Prolific platform. For the pre-screening survey, we used the same language proficiency questions as in the pre-task survey (Appendix \ref{appendix:design_details}) to reliably recruit monolingual English speakers. To ensure high-quality responses, we limited participation to users with a Prolific approval rate above 90\% and at least 10 prior submissions. We recruited 205 participants and invited 123 who selected ``I can read long and difficult texts in English without help'' for English proficiency, and either ``I cannot understand any Spanish words or sentences at all'' or ``I can read some Spanish words and very simple sentences, such as greetings and common expressions'' for Spanish proficiency. Each pre-screening participant received 0.20 USD, totaling 55.20 USD including Prolific platform fees.

\paragraph{Main task.}
Of the 123 invited participants, 91 completed the main task. Each received a base payment of 5 USD for 20 minutes of participation, with an additional 2 USD performance-based bonus for those achieving over 70\% overall decision accuracy. A total of 30 participants qualified for this bonus. One participant who failed both attention check questions was compensated but excluded from the final analysis, resulting in 90 valid responses. The median task completion time was 14 minutes, corresponding to an effective pay rate of 20 USD per hour. The total cost for the main task, including Prolific fees, was 588.01 USD.

\section{Inferring AI Feedback Decisions}
\label{appendix:error_rate}

During dataset selection, we ensure that each feedback condition includes a relatively balanced number of examples reflecting both correct and incorrect AI predictions. Since none of the feedback types provide a direct prediction of shareability judgment, we apply  tailored proxy measures to approximate the intended shareability label for each feedback type, as detailed below.

\paragraph{Error Highlights.}
For each error annotation, we take the highest error severity level h (No error, Minor, Major, or Critical) as the QE model’s decision. The decision is mapped to the two-point shareability scale as follows:
\begin{equation}
\small
\begin{cases}
    \text{\raisebox{-0.2em}{\includegraphics[height=1.1em]{figures/good.png}} Safe to share as-is}, & h \in \{\text{No error}, \text{Minor}\} \\
    \text{\raisebox{-0.2em}{\includegraphics[height=1.1em]{figures/bad.png}} Needs bilingual review}, & h \in \{\text{Major}, \text{Critical}\}
\end{cases}
\end{equation}

\paragraph{LLM Explanation.}
We prompt the same model used to generate textual explanations (\textsc{LLaMA-3.3 70B}) to make shareability judgments using the same information and option format provided to participants in the task instructions. Exact prompt used is outlined in Appendix~\ref{appendix:prompt_explanation}.

\paragraph{Backtranslation.}
Following \citet{ki2025askqequestionansweringautomatic}, we compute \textsc{xCOMET-QE} scores between the Spanish MT and its backtranslation. We fit a three-component Gaussian Mixture Model (GMM).\footnote{\url{https://scikit-learn.org/stable/modules/mixture.html}} GMM is a probabilistic clustering model that assumes data points are generated from a mixture of Gaussian distributions, assigning a probability to each point for belonging to a specific cluster. We hypothesize two clusters corresponding to different ranges of QE scores: \textbf{(1)} \raisebox{-0.2em}{\includegraphics[height=1.1em]{figures/good.png}} ``Safe to share as-is'' cluster with higher QE scores and \textbf{(2)} \raisebox{-0.2em}{\includegraphics[height=1.1em]{figures/bad.png}} ``Needs bilingual review before sharing'' cluster with lower QE scores. Each of the 20 examples is assigned a probability to each cluster and is categorized into the cluster with the highest probability.

\paragraph{QA Table.}
We follow a similar procedure to that used for backtranslation. However, instead of using QE scores, we compute \textsc{AskQE} scores through a two-step process: (1) measuring overlap between answers from the source $A_{\mathrm{src}}$ and from the backtranslated MT output $A_{\mathrm{bt}}$, and (2) aggregating question-answer similarities into a segment-level metric. We use \textsc{F1}, a standard similarity measure in QA research \citep{rajpurkar-etal-2016-squad, deutsch-etal-2021-towards}. Formally, for each example $e \in E$, we compute:
\begin{equation}
    \text{AskQE}(e) = \sum_{i=1}^{E}{\frac{\mathrm{F1}(A_{\mathrm{src}}, A_{\mathrm{bt}})}{E}}
\end{equation}
We fit a two-component Gaussian Mixture Model (GMM) to the \textsc{AskQE} scores and assign each example to the cluster with the highest probability.

The selected examples were then manually reviewed by the authors to ensure that the proxy measures provided a reasonable approximation of the feedback’s implied shareability judgments. This process resulted in a relatively balanced distribution of correct and incorrect examples across feedback types: we included 9, 10, 11, and 9 correct examples for the Error Highlights, LLM Explanation, Backtranslation, and QA Table conditions, respectively (with corresponding incorrect counts of 11, 10, 9, and 10). This helped ensure that each feedback condition reflected a comparable underlying error rate.

\section{Computational \& Time Efficiency}
\label{appendix:efficiency}
We compare the average computational and time efficiency of generating four types of quality feedback across 20 examples chosen from the gold annotation process. As shown in Table~\ref{tab:efficiency}, all feedback demonstrate sufficient efficiency for deployment in user-facing applications.

\begin{table}
\centering
\resizebox{\linewidth}{!}{%
    \begin{tabular}{llllllll}
    \specialrule{1.3pt}{0pt}{0pt}
    \textbf{Feedback} & \textbf{Model (\textit{size})} & \textbf{Computation} & \textbf{Avg. Time} \\
    \toprule

    \textbf{Error Highlights} & \textsc{xCOMET-XXL} (10.7B) & $\times$ 3 & 03:52 \\
    \textbf{LLM Explanation} & \textsc{LLaMA-3.3} (70B) & $\times$ 8 & 01:56 \\
    \textbf{Backtranslation} & Google Translate & - & 01:24 \\
    \textbf{QA Table} & \textsc{LLaMA-3.3} (70B) & $\times$ 8 & 03:48 \\
    
    \specialrule{1.3pt}{0pt}{0pt}
    \end{tabular}
}
\caption{Average computational and time efficiency for four types of quality feedback. \textbf{Computation} is reported in GPU units (RTX A5000). \textbf{Avg. Time} is shown in MM:SS.} 
\label{tab:efficiency}
\end{table}

\section{Detailed Results}
\label{appendix:detailed_results}

\subsection{Independent vs. AI-Assisted}
\label{appendix:independent_aiassisted}
We present detailed results for both the independent and AI-assisted decision-making steps in terms of decision accuracy and \textsc{CWA} in Table \ref{tab:detailed_results}. We show that the QA table feedback yields the highest overall and AI-assisted performance across both metrics, while error highlights have the lowest scores.

\subsection{Switch Percentage}
\label{appendix:detailed_switch}
To account for the differing consequences of reliance relative to shareability, we first define four outcome types based on the participant’s shareability judgment $\hat{s}$ and the gold label $s^*$:
\begin{itemize}[leftmargin=*, itemsep=2pt, parsep=-1pt]
 \item \textbf{True Positive (TP)}: $s^*$ is shareable, $\hat{s} = s^*$
 \item \textbf{False Positive (FP)}: $s^*$ is shareable, $\hat{s} \neq s^*$
  \item \textbf{True Negative (TN)}: $s^*$ is non-shareable, $\hat{s} = s^*$
 \item \textbf{False Negative (FN)}: $s^*$ is non-shareable, $\hat{s}\neq s^*$
\end{itemize}
Using this, we categorize decision transitions as shown in Table~\ref{tab:transition}: \textbf{(1)} Over-reliance: changing from a correct to an incorrect decision after feedback (TP $\rightarrow$ FP and TN $\rightarrow$ FN); \textbf{(2)} Under-reliance: maintaining an incorrect decision after feedback (FP $\rightarrow$ FP and FN $\rightarrow$ FN); \textbf{(3)} Appropriate reliance: either maintaining a correct decision (TP $\rightarrow$ TP and TN $\rightarrow$ TN) or correcting an incorrect one after feedback (FP $\rightarrow$ TP and FN $\rightarrow$ TN).

\begin{table}[htbp]
\centering
\resizebox{\linewidth}{!}{%
    \begin{tabular}{llllllll}
    \specialrule{1.3pt}{0pt}{0pt}
    \textbf{} & \textbf{TP} & \textbf{FP} & \textbf{TN} & \textbf{FN} \\
    \toprule

    \textbf{TP} & Appropriate & Over & - & - \\
    \textbf{FP} & Appropriate & Under & - & - \\
    \textbf{TN} & - & - & Appropriate & Over \\
    \textbf{FN} & - & - & Appropriate & Under \\

    \specialrule{1.3pt}{0pt}{0pt}
    \end{tabular}
}
\caption{Categorization of decision transitions from Independent (\textit{rows}) to AI-Assisted (\textit{columns}). \textbf{Appropriate:} Appropriate reliance; \textbf{Over:} Over-reliance; \textbf{Under:} Under-reliance.} 
\label{tab:transition}
\end{table}

In Table \ref{tab:detailed_switch}, we present detailed switch percentage results for each quality feedback type, broken down into each constituent of under-, over-, and appropriate reliance. Across all conditions, participants show higher over-reliance and lower under-reliance for shareable examples than non-shareable ones.

\begin{table}[htbp]
\centering
\resizebox{\linewidth}{!}{%
    \begin{tabular}{llllllll}
    \specialrule{1.3pt}{0pt}{0pt}
    \textbf{Feedback} & \textbf{Reliance} & \textbf{Transition} & \textbf{Value (\%)} \\
    \toprule

    \multirow{8}{*}{\textbf{Error Highlights}} 
    & \multirow{4}{*}{Appropriate reliance} & TP $\rightarrow$ TP & 113 (31.4) \\
    & & FP $\rightarrow$ TP & 23 (6.39) \\
    & & TN $\rightarrow$ TN & 67 (18.6) \\
    & & FN $\rightarrow$ TN & 29 (8.06) \\
    \cmidrule{2-4}
    & \multirow{2}{*}{Over-reliance} & TP $\rightarrow$ FP & 20 (5.56) \\
    & & TN $\rightarrow$ FN & 16 (4.44) \\
    \cmidrule{2-4}
    & \multirow{2}{*}{Under-reliance} & FP $\rightarrow$ FP & 26 (7.22) \\
    & & FN $\rightarrow$ FN & 66 (18.3) \\
    \midrule

    \multirow{8}{*}{\textbf{LLM Explanation}} & \multirow{4}{*}{Appropriate reliance} & TP $\rightarrow$ TP & 100 (27.8) \\
    & & FP $\rightarrow$ TP & 24 (6.67) \\
    & & TN $\rightarrow$ TN & 78 (21.7) \\
    & & FN $\rightarrow$ TN & 45 (12.5) \\
    \cmidrule{2-4}
    & \multirow{2}{*}{Over-reliance} & TP $\rightarrow$ FP & 31 (8.61) \\
    & & TN $\rightarrow$ FN & 17 (4.72) \\
    \cmidrule{2-4}
    & \multirow{2}{*}{Under-reliance} & FP $\rightarrow$ FP & 25 (6.94) \\
    & & FN $\rightarrow$ FN & 40 (11.1) \\
    \midrule

    \multirow{8}{*}{\textbf{Backtranslation}} & \multirow{4}{*}{Appropriate reliance} & TP $\rightarrow$ TP & 115 (31.9) \\
    & & FP $\rightarrow$ TP & 31 (8.61) \\
    & & TN $\rightarrow$ TN & 75 (20.8) \\
    & & FN $\rightarrow$ TN & 37 (10.3) \\
    \cmidrule{2-4}
    & \multirow{2}{*}{Over-reliance} & TP $\rightarrow$ FP & 10 (2.78) \\
    & & TN $\rightarrow$ FN & 16 (4.44) \\
    \cmidrule{2-4}
    & \multirow{2}{*}{Under-reliance} & FP $\rightarrow$ FP & 24 (6.67) \\
    & & FN $\rightarrow$ FN & 52 (14.4) \\
    \midrule

    \multirow{8}{*}{\textbf{QA Table}} & \multirow{4}{*}{Appropriate reliance} & TP $\rightarrow$ TP & 123 (34.2) \\
    & & FP $\rightarrow$ TP & 29 (8.06) \\
    & & TN $\rightarrow$ TN & 86 (23.9) \\
    & & FN $\rightarrow$ TN & 39 (10.8) \\
    \cmidrule{2-4}
    & \multirow{2}{*}{Over-reliance} & TP $\rightarrow$ FP & 16 (4.44) \\
    & & TN $\rightarrow$ FN & 12 (3.33) \\
    \cmidrule{2-4}
    & \multirow{2}{*}{Under-reliance} & FP $\rightarrow$ FP & 11 (3.06) \\
    & & FN $\rightarrow$ FN & 44 (12.2) \\
    
    \specialrule{1.3pt}{0pt}{0pt}
    \end{tabular}
}
\caption{Detailed switch percentage results by quality feedback type and reliance (appropriate, over-, under-reliance).} 
\label{tab:detailed_switch}
\end{table}

\subsection{Per Shareability Label}
\label{appendix:shareability_label}
We show detailed decision accuracy and \textsc{CWA} scores by shareability label (\raisebox{-0.2em}{\includegraphics[height=1.1em]{figures/good.png}} Safe to share as-is and \raisebox{-0.2em}{\includegraphics[height=1.1em]{figures/bad.png}} Needs bilingual review before sharing) in Table \ref{tab:detailed_label}. Across all conditions, both decision accuracy and \textsc{CWA} are consistently higher for examples labeled as shareable, indicating that participants make more accurate and better-calibrated decisions for good translations than for those requiring bilingual review.

\subsection{Participant Responses}
\label{appendix:detailed_comments}
We present detailed participants' comments of what aspects of the quality feedback they received to be helpful or unhelpful. This question varied depending on their rating of helpfulness, as shown in Appendix \ref{appendix:design_details}.

\begin{table*}
\centering
\resizebox{\linewidth}{!}{%
    \begin{tabular}{llllllll}
    \specialrule{1.3pt}{0pt}{0pt}
    \textbf{Feedback} & \textbf{Decision Acc. (Total)} & \textbf{Decision Acc. (Ind.)} & \textbf{Decision Acc. (AI)} & \textbf{\textsc{CWA} (Total)} & \textbf{\textsc{CWA} (Ind.)} & \textbf{\textsc{CWA} (AI)} \\
    \toprule

    \textbf{Control} & 0.609 \small{[0.558, 0.660]} & 0.613 \small{[0.542, 0.684]} & 0.605 \small{[0.533, 0.678]} & 0.177 \small{[0.101, 0.253]} & 0.182 \small{[0.074, 0.290]} & 0.172 \small{[0.066, 0.278]} \\

    \textbf{Error Highlights} & 0.620 \small{[0.547, 0.693]} & 0.593 \small{[0.518, 0.668]} & 0.647 \small{[0.565, 0.730]} & 0.197 \small{[0.063, 0.331]} & 0.156 \small{[0.020, 0.292]} & 0.239 \small{[0.089, 0.388]} \\
    
    \textbf{LLM Explanation} & 0.664 \small{[0.610, 0.717]} & 0.634 \small{[0.567, 0.701]} & 0.693 \small{[0.631, 0.756]} & 0.292 \small{[0.193, 0.392]} & 0.232 \small{[0.111, 0.352]} & 0.353 \small{[0.241, 0.465]} \\

    \textbf{Backtranslation} & 0.659 \small{[0.603, 0.715]} & 0.603 \small{[0.540, 0.666]} & 0.716 \small{[0.683, 0.847]} & 0.288 \small{[0.188, 0.389]} & 0.188 \small{[0.079, 0.298]} & 0.388 \small{[0.272, 0.505]} \\
    
    \textbf{QA Table} & \textbf{0.712} \small{[0.639, 0.786]} & \textbf{0.659} \small{[0.573, 0.746]} & \textbf{0.765} \small{[0.683, 0.847]} & \textbf{0.385} \small{[0.252, 0.519]} & \textbf{0.282} \small{[0.131, 0.433]} & \textbf{0.488} \small{[0.333, 0.643]} \\
    
    \specialrule{1.3pt}{0pt}{0pt}
    \end{tabular}
}
\caption{Detailed results for decision accuracy and \textsc{CWA} by condition. \textbf{Ind.:} Independent decision-making step; \textbf{AI:} AI-assisted decision-making step. Values represent mean scores with corresponding 95\% confidence intervals. Best scores for each column is \textbf{bold}.} 
\label{tab:detailed_results}
\end{table*}
\begin{table*}
\centering
\resizebox{\linewidth}{!}{%
    \begin{tabular}{llllllll}
    \specialrule{1.3pt}{0pt}{0pt}
    \textbf{Step} & \textbf{Feedback} & \textbf{Decsion Acc.} (\raisebox{-0.2em}{\includegraphics[height=1em]{figures/good.png}}) & \textbf{Decision Acc.} (\raisebox{-0.2em}{\includegraphics[height=1em]{figures/bad.png}}) & \textbf{\textsc{CWA}} (\raisebox{-0.2em}{\includegraphics[height=1em]{figures/good.png}}) & \textbf{\textsc{CWA}} (\raisebox{-0.2em}{\includegraphics[height=1em]{figures/bad.png}}) \\
    \toprule

    \textbf{Independent} & - & 0.733 \small{[0.701, 0.765]} & 0.509 \small{[0.473, 0.546]} & 0.402 \small{[0.351, 0.453]} & 0.024 \small{[-0.034, 0.081]} \\
    \midrule
    
    \multirow{4}{*}{\textbf{AI-Assisted}}     
    & \textbf{Error Highlights} & 0.753 \small{[0.690, 0.816]} & 0.546 \small{[0.474, 0.619]} & 0.415 \small{[0.310, 0.520]} & 0.069 \small{[-0.051, 0.189]} \\
    
    & \textbf{LLM Explanation} & 0.709 \small{[0.644, 0.774]} & 0.681 \small{[0.614, 0.748]} & 0.380 \small{[0.266, 0.494]} & 0.323 \small{[0.204, 0.443]} \\

    & \textbf{Backtranslation} & 0.805 \small{[0.746, 0.864]} & 0.622 \small{[0.549, 0.695]} & 0.566 \small{[0.467, 0.664]} & 0.210 \small{[0.082, 0.339]} \\
    
    & \textbf{QA Table} & \textbf{0.848} \small{[0.795, 0.901]} & \textbf{0.687} \small{[0.619, 0.755]} & \textbf{0.649} \small{[0.554, 0.745]} & \textbf{0.335} \small{[0.217, 0.455]} \\
    
    \specialrule{1.3pt}{0pt}{0pt}
    \end{tabular}
}
\caption{Detailed results for decision accuracy and \textsc{CWA} by quality feedback type and shareability label. \raisebox{-0.2em}{\includegraphics[height=1.1em]{figures/good.png}}: Safe to share as-is; \raisebox{-0.2em}{\includegraphics[height=1.1em]{figures/bad.png}}: Needs bilingual review before sharing. Values represent mean scores with corresponding 95\% confidence intervals. Best scores for each column is \textbf{bold}.} 
\label{tab:detailed_label}
\end{table*}

\begin{table*}
\centering
\resizebox{\linewidth}{!}{%
    \begin{tabular}{c p{21cm}}
    \specialrule{1.3pt}{0pt}{0pt}
    \textbf{Feedback} & \textbf{Participant Responses} \\
    \toprule

    \multirow{8}{*}{\raisebox{-0.2em}{\includegraphics[height=1.1em]{figures/logos/annotation.png}}} & \textbullet~The fact that it doesn't tell me what the highlighted translated words mean makes it useless. When it tells me certain things are translated incorrectly but doesn't tell me what the meaning of the translation is, that's so pointless. (\raisebox{-0.2em}{\includegraphics[height=1.1em]{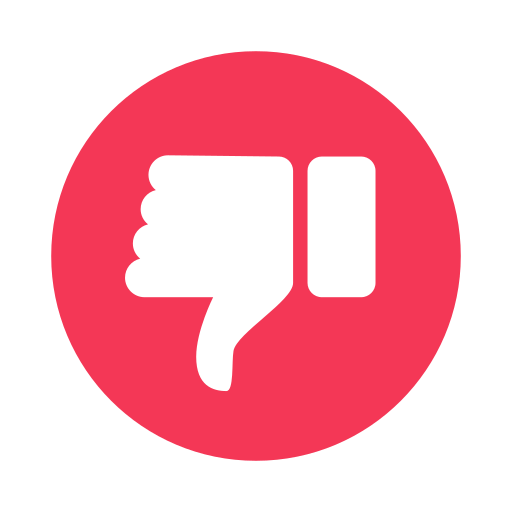}}) \\
    & \textbullet~Since I \textbf{don't understand Spanish} I can't even judge how wrong/right the error detection is. I can't see this being helpful to anyone except people who already know Spanish. (\raisebox{-0.2em}{\includegraphics[height=1.1em]{tables/down.png}}) \\
    & \textbullet~It gave me the suggestions about the information \textbf{I could not understand}. (\raisebox{-0.2em}{\includegraphics[height=1.1em]{tables/down.png}}) \\

    & \textbullet~It helped highlight \textbf{areas that not have been accurately translated}. (\raisebox{-0.2em}{\includegraphics[height=1.1em]{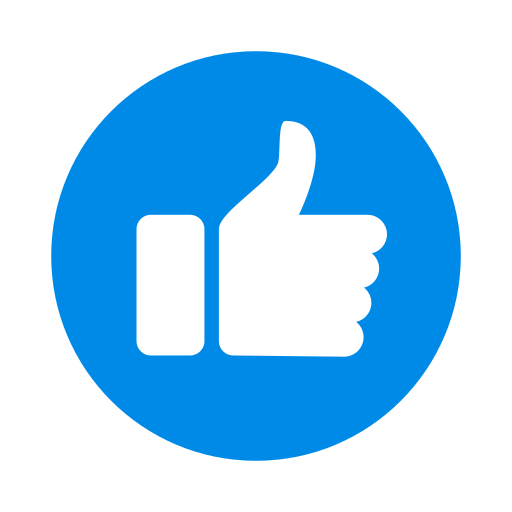}}) \\
    & \textbullet~It let me know where to look, and if it thought there were errors. (\raisebox{-0.2em}{\includegraphics[height=1.1em]{tables/up.png}}) \\
    & \textbullet~It highlighted \textbf{key translation mistakes}, helping to ensure the translation was accurate before sharing. (\raisebox{-0.2em}{\includegraphics[height=1.1em]{tables/up.png}}) \\
    \midrule

    \multirow{6}{*}{\raisebox{-0.2em}{\includegraphics[height=1.1em]{figures/logos/explanation.png}}} 
    & \textbullet~Provided \textbf{exactly} why it was incorrect or correct. (\raisebox{-0.2em}{\includegraphics[height=1.1em]{tables/up.png}}) \\
    & \textbullet~\textbf{Gave me the answers.} I don't know Spanish. (\raisebox{-0.2em}{\includegraphics[height=1.1em]{tables/up.png}}) \\
    & \textbullet~It has enabled to know whether the translation was accurate or not. (\raisebox{-0.2em}{\includegraphics[height=1.1em]{tables/up.png}}) \\
    & \textbullet~The information provided by explanation helped me \textbf{understand the context of what was being said} more clearly. (\raisebox{-0.2em}{\includegraphics[height=1.1em]{tables/up.png}}) \\
    & \textbullet~By explaining what was wrong with the translation and putting what would be the \textbf{correct translation}. (\raisebox{-0.2em}{\includegraphics[height=1.1em]{tables/up.png}}) \\
    & \textbullet~It clarified some phrasing that was clearly a \textbf{direct translation versus a natural translation}. (\raisebox{-0.2em}{\includegraphics[height=1.1em]{tables/up.png}}) \\
    \midrule

    \multirow{6}{*}{\raisebox{-0.2em}{\includegraphics[height=1.1em]{figures/logos/bt.png}}} 
    & \textbullet~Since I'm not very fluent in Spanish, the AI-translation was helpful in \textbf{verifying some tiny unsure details}.(\raisebox{-0.2em}{\includegraphics[height=1.1em]{tables/up.png}}) \\
    & \textbullet~I made a variety of mistakes when attempting to translate myself, the AI was helpful in translating particular words which made a \textbf{major difference in the meaning} of the sentence. (\raisebox{-0.2em}{\includegraphics[height=1.1em]{tables/up.png}}) \\
    & \textbullet~It provides insight into whether the \textbf{core meaning and intent of the original text are preserved} in the translation. (\raisebox{-0.2em}{\includegraphics[height=1.1em]{tables/up.png}}) \\
    & \textbullet~The information \textbf{simplified} what was said and made it very easy to understand. (\raisebox{-0.2em}{\includegraphics[height=1.1em]{tables/up.png}}) \\
    & \textbullet~Since I'm not very fluent in Spanish, the AI translation was helpful in verifying some tiny unsure details. (\raisebox{-0.2em}{\includegraphics[height=1.1em]{tables/up.png}}) \\
    \midrule

    \multirow{9}{*}{\raisebox{-0.2em}{\includegraphics[height=1.1em]{figures/logos/qa.png}}} 
    & \textbullet~If the statement is blue, I \textbf{double check} the phrase again. (\raisebox{-0.2em}{\includegraphics[height=1.1em]{tables/up.png}}) \\
    & \textbullet~I could compare the words and \textbf{make determinations}. (\raisebox{-0.2em}{\includegraphics[height=1.1em]{tables/up.png}}) \\
    & \textbullet~It was helpful in terms of giving a \textbf{breakdown} of the Spanish translation. (\raisebox{-0.2em}{\includegraphics[height=1.1em]{tables/up.png}}) \\
    & \textbullet~It provided some clarity in showing that there was a \textbf{difference in the translation}. (\raisebox{-0.2em}{\includegraphics[height=1.1em]{tables/up.png}}) \\
    & \textbullet~I liked seeing the orange and blue so I knew if similar or different. These helped when \textbf{slight differences} such as when things were mild or the other said severe. (\raisebox{-0.2em}{\includegraphics[height=1.1em]{tables/up.png}}) \\
    & \textbullet~There were \textbf{small but seriously important words} I missed that when giving advice on an issue that could mean life or death matter. (\raisebox{-0.2em}{\includegraphics[height=1.1em]{tables/up.png}}) \\
    & \textbullet~It helped me know what information was being displayed on the Spanish side. (\raisebox{-0.2em}{\includegraphics[height=1.1em]{tables/up.png}}) \\
    & \textbullet~I would like this to be included for future surveys. (\raisebox{-0.2em}{\includegraphics[height=1.1em]{tables/up.png}}) \\

    \specialrule{1.3pt}{0pt}{0pt}
    \end{tabular}
}
\caption{Participants' responses on how they used quality feedback in their decision-making process, specifically why the feedback was perceived as confusing or unhelpful (\raisebox{-0.2em}{\includegraphics[height=1.1em]{tables/down.png}}) or why it was considered helpful (\raisebox{-0.2em}{\includegraphics[height=1.1em]{tables/up.png}}).}
\label{tab:comments}
\end{table*}

\begin{figure*}
    \setcounter{subfigure}{7}
    \centering
    \subfigure[AI-Assisted: \raisebox{-0.2em}{\includegraphics[height=1.1em]{figures/logos/bt.png}} Backtranslation]{\includegraphics[width=0.49\textwidth]{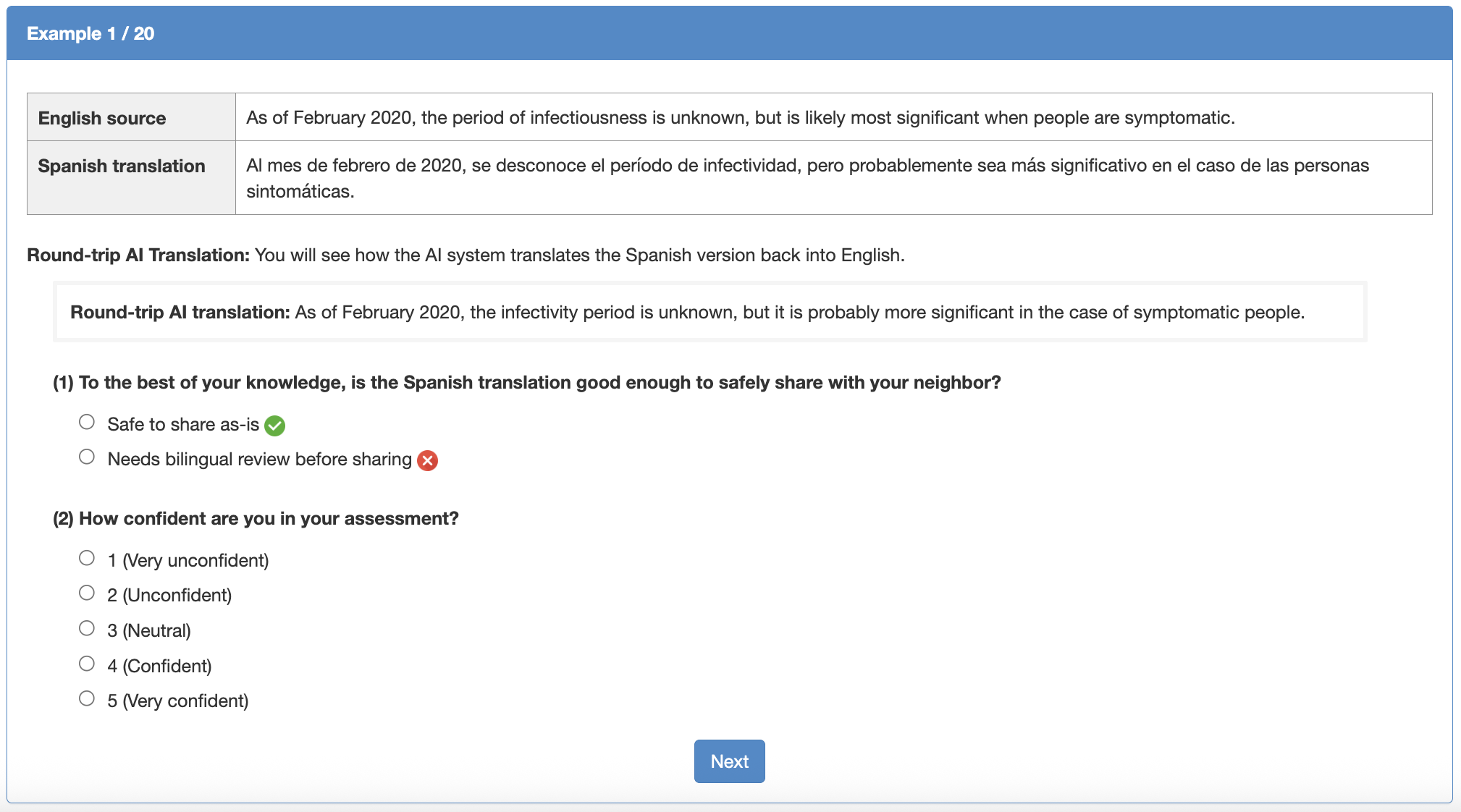}}
    \subfigure[AI-Assisted: \raisebox{-0.2em}{\includegraphics[height=1.1em]{figures/logos/qa.png}} QA Table]{\includegraphics[width=0.49\textwidth]{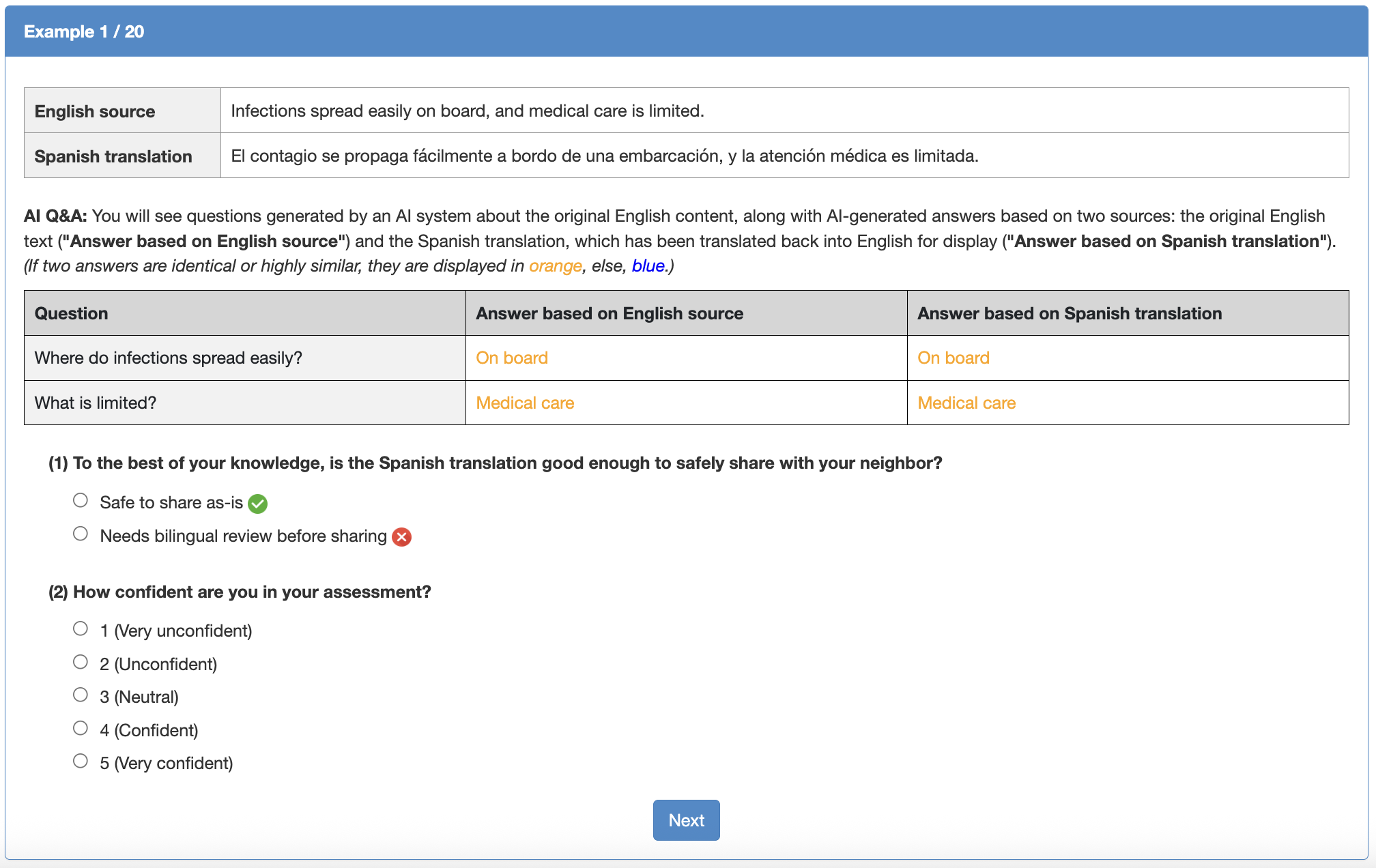}} 
    
    \subfigure[Attention Check Question]{\includegraphics[width=0.49\textwidth]{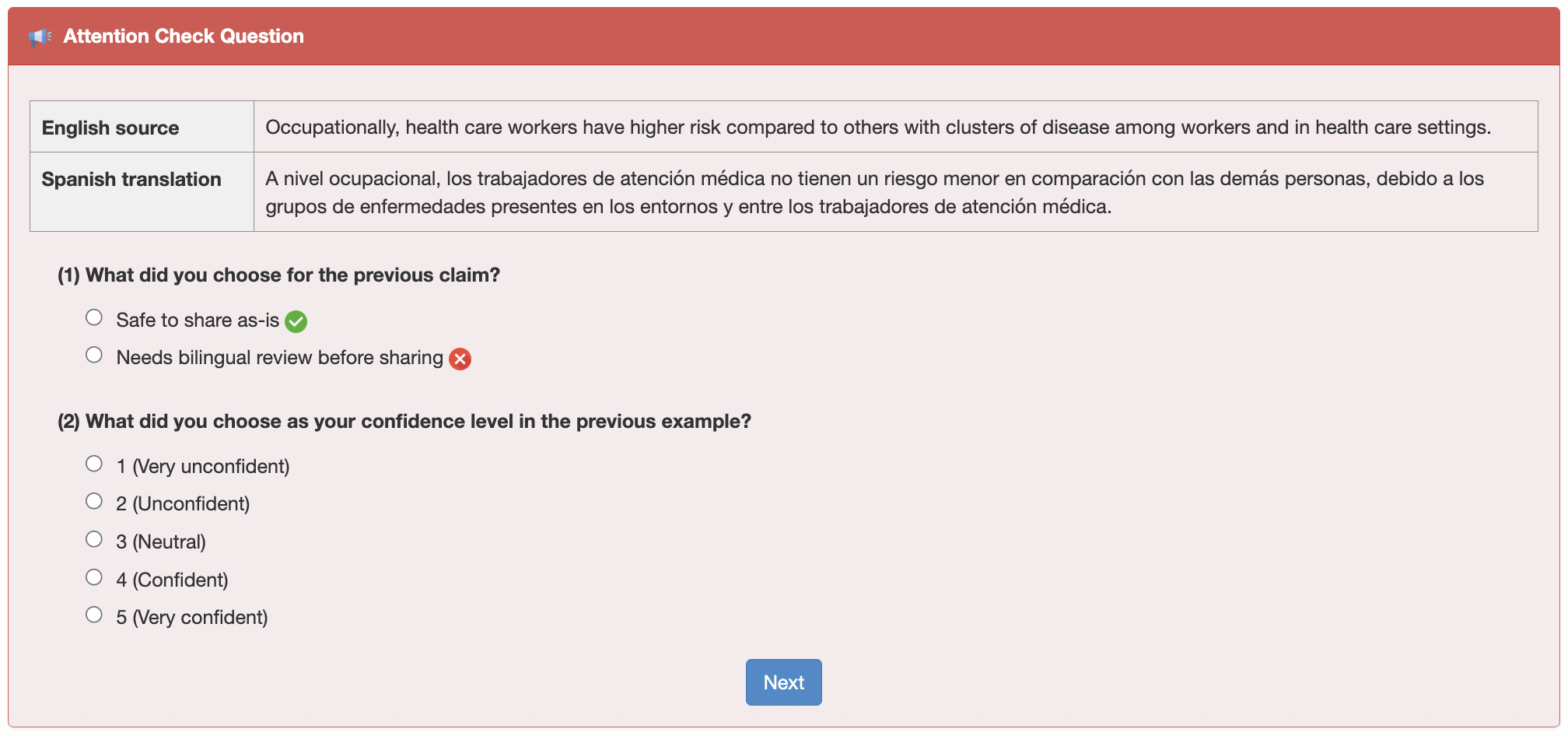}}
    \subfigure[Post-task Survey]{\includegraphics[width=0.49\textwidth]{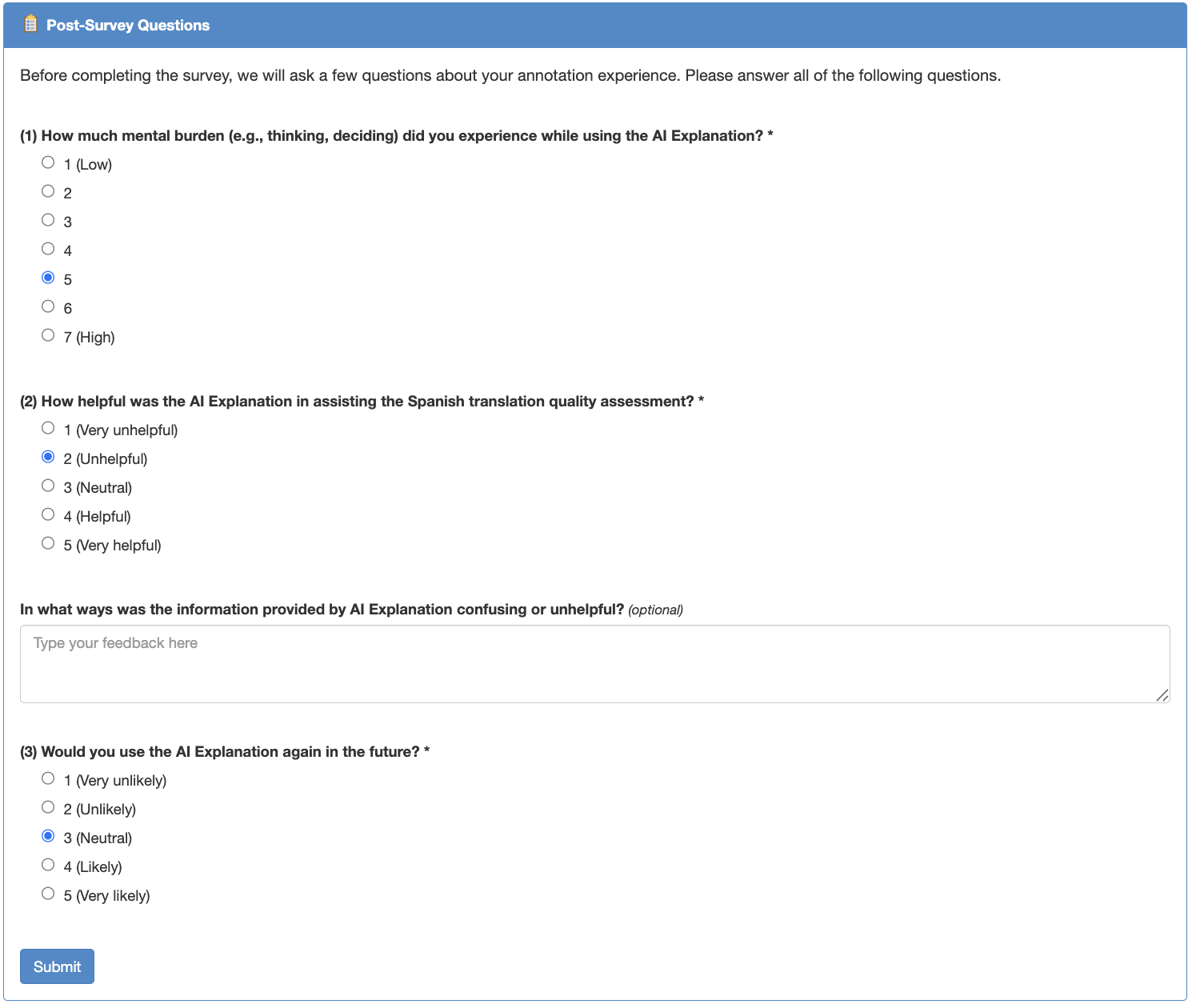}} 
\end{figure*}

\end{document}